%% file: main.tex
\documentclass{article}

\title{Systems and Algorithms for Convolutional\\ Multi-Hybrid Language Models at Scale}

\author{%
  Jerome Ku$^{2,*}$, 
  Eric Nguyen$^{1,*}$, 
  David W. Romero$^{3,*}$, 
  Garyk Brixi$^{1}$, 
  Brandon Yang$^{5}$, \\
  Anton Vorontsov$^{3}$, 
  Ali Taghibakhshi$^{3}$, 
  Amy X. Lu$^{4}$, 
  Dave P. Burke$^{2}$, \\
  Greg Brockman$^{5,\dagger}$, 
  Stefano Massaroli$^{7, 8}$, 
  Christopher Ré$^{1}$, 
  Patrick D. Hsu$^{2,4}$, \\
  Brian L. Hie$^{1,2}$, 
  Stefano Ermon$^{1}$, 
  Michael Poli$^{1,7,\ddagger}$
}
\date{\small
  $^1$Stanford University,
  $^2$Arc Institute,
  $^3$NVIDIA, \\
  $^4$University of California, Berkeley,
  $^5$Independent Researcher,
  $^7$Liquid AI,
  $^8$RIKEN
}
\input{template/new_style}
\input{utils/math}
\begin{document}

\maketitle

\begin{abstract}
    \noindent
    We introduce\let\thefootnote\relax\footnotetext{\hspace{-0.6cm}$^*$ These authors contributed equally to this work. \\$^\dagger$ Current address: OpenAI. \\$^{\ddagger}$Corresponding author: poli@stanford.edu.} \textit{convolutional multi-hybrid} architectures, with a design grounded on two simple observations. First, operators in hybrid models can be tailored to token manipulation tasks such as in-context recall, multi-token recall, and compression, with input-dependent convolutions and attention offering complementary performance. Second, co-designing convolution operators and hardware-aware algorithms enables efficiency gains in regimes where previous alternative architectures struggle to surpass Transformers. At the 40 billion parameter scale, we train end-to-end 1.2 to 2.9 times faster than optimized Transformers, and 1.1 to 1.4 times faster than previous generation hybrids. On H100 GPUs and model width 4096, individual operators in the proposed multi-hybrid StripedHyena 2 architecture achieve two-fold throughput improvement over linear attention and state-space models. Multi-hybrids excel at sequence modeling over byte-tokenized data, as demonstrated by the Evo 2 line of models. We discuss the foundations that enable these results, including architecture design, overlap-add blocked kernels for tensor cores, and dedicated all-to-all and point-to-point context parallelism strategies.  
\end{abstract}

\input{sections/1_introduction}

\input{sections/architecture.tex}

\input{sections/kernel_optimizations.tex}

\input{sections/context_parallelism.tex}

\input{sections/discussion.tex}

\bibliography{main}
\bibliographystyle{style}

\newpage
\clearpage

\appendix

\rule[0pt]{\columnwidth}{1pt}
\begin{center}
    \Large{Systems and Algorithms for Convolutional \\ Multi-Hybrid Language Models at Scale} \\
    \vspace{0.15cm}
    \emph{Supplementary Material}
\end{center}
\rule[0pt]{\columnwidth}{1.2pt}

\doparttoc
\tableofcontents

\input{sections/appendix/A_details}
\input{sections/appendix/B_figures}
\input{sections/appendix/C_protocol}

\end{document}

%% file: template/new_style.tex
\usepackage[T1]{fontenc}  
\usepackage[bookmarks=true, 
            colorlinks=true,
            linkcolor=bluegray,
            citecolor=junglegreen]{hyperref}
\usepackage{natbib}
\usepackage[a4paper, margin=.9in]{geometry}
\usepackage{wrapfig}
\usepackage{booktabs}  
\usepackage[bottom]{footmisc}
\usepackage{enumitem} 

\usepackage{chngcntr}
\counterwithin{figure}{section} 
\counterwithin{table}{section}

\usepackage{graphicx}
\usepackage{tikz}
\usepackage{tikz-cd}
\usepackage{hf-tikz}
\usepackage{pgfplots} 
\pgfplotsset{compat=1.17} 
\pgfplotsset{
        table/search path={figures/drawings},
    }
\usetikzlibrary{fadings}
\usetikzlibrary{shapes, arrows, fit, backgrounds, arrows.meta}

\usetikzlibrary{matrix}
\usetikzlibrary{shadows.blur}
\usetikzlibrary{patterns, tikzmark}
\usetikzlibrary{decorations.pathreplacing, calc, decorations.markings,}
\usetikzlibrary{positioning}

\usepgfplotslibrary{groupplots}
\usepgfplotslibrary{patchplots}
\definecolor{bluegray}{rgb}{0.4, 0.6, 0.8}
\definecolor{bluebell}{rgb}{0.64, 0.64, 0.82}
\definecolor{etonblue}{rgb}{0.59, 0.78, 0.64}
\definecolor{junglegreen}{rgb}{0.16, 0.67, 0.53}
\definecolor{bg}{gray}{0.97}
\definecolor{olive}{rgb}{0.6, 0.6, 0.2}
\definecolor{sand}{rgb}{0.8666666666666667, 0.8, 0.4666666666666667}
\definecolor{wine}{rgb}{0.5333333333333333, 0.13333333333333333, 0.3333333333333333}
\definecolor{deblue}{RGB}{11,132,147}
\definecolor{ocra}{RGB}{204, 119, 34}

\pgfplotsset{%
            mesh line legend/.style={legend image code/.code=\meshlinelegend#1},%
}
\makeatletter
\long\def\meshlinelegend#1{%
    \scope[%
        #1,
        /pgfplots/mesh/rows=1,
        /pgfplots/mesh/cols=4,
        /pgfplots/mesh/num points=,
        /tikz/x={(0.44237cm,0cm)}, 
        /tikz/y={(0cm,0.23932cm)},
        /tikz/z={(0.0cm,0cm)},
        scale=0.4,
    ]
    \let\pgfplots@metamax=\pgfutil@empty
    \pgfplots@curplot@threedimtrue

    \pgfplotsplothandlermesh
    \pgfplotstreamstart

    \def\simplecoordinate(##1,##2,##3){%
        \pgfmathparse{1000*(##3)}%
        \pgfmathfloatparsenumber\pgfmathresult
        \let\pgfplots@current@point@meta=\pgfmathresult
        \pgfplotstreampoint{\pgfqpointxyz@orig{##1}{##2}{##3}}%
    }%

    \pgfplotsforeachungrouped \x in {0,...,\pgfkeysvalueof{/pgfplots/samples}}{
        \pgfmathsetmacro\y{\x/\pgfkeysvalueof{/pgfplots/samples}}
        \pgfmathsetmacro\x{\x/\pgfkeysvalueof{/pgfplots/samples}*3}
        \simplecoordinate(\x,0,\y)
    }

    \pgfplotstreamend
    \pgfusepath{stroke}
    \endscope
}%
\makeatother

\usepackage{lettrine}
\usepackage{Typocaps}

\LettrineTextFont{\itshape}
\usepackage{amsthm}

\usepackage{minitoc}

\newcommand{\chapref}[1]{\hyperref[#1]{Chapter \ref{#1}}}
\newcommand{\secref}[1]{\hyperref[#1]{Section \ref{#1}}}

\usepackage{marginnote}

\usepackage[many]{tcolorbox}
\tcbuselibrary{breakable,xparse,skins}
\usepackage{bigstrut}
\usepackage{newunicodechar}
\newunicodechar{λ}{{$\mathtt\lambda$}}
\newunicodechar{μ}{{$\mathtt\mu$}}
\newtcolorbox{note}[1]{breakable, enhanced, sharp corners, frame hidden, #1}

\usepackage{newunicodechar}
\newunicodechar{λ}{{$\mathtt\lambda$}}
\newunicodechar{μ}{{$\mathtt\mu$}}

\usepackage{xparse}
\DeclareTColorBox{emphbox}{O{black}O{0cm}}{
    enhanced jigsaw,
    breakable,
    outer arc=0pt,
    arc=0pt,
    colback=white,
    rightrule=0pt,
    toprule=0pt,
    top=0pt,
    right=0pt,
    bottom=0pt,
    bottomrule=0pt,
    colframe=#1,
    enlarge left by=#2,
    width=\linewidth-#2,
}

\DeclareTColorBox{emphbox}{O{black}O{0cm}}{
    empty,
    breakable=true,
    outer arc=0pt,
    arc=0pt,
    rightrule=0pt,
    leftrule=2pt,
    borderline west={2pt}{0pt}{#1},
    toprule=0pt,
    top=0pt,
    right=-3pt,
    bottom=0pt,
    bottomrule=0pt,
    colframe=#1,
    enlarge left by=#2,
    width=\linewidth-#2,
}
\usepackage{listings}
\makeatletter
\newcommand\BeraMonottfamily{%
  \def\fvm@Scale{0.85}
  \fontfamily{fvm}\selectfont
}
\makeatother

\usepackage[parfill]{parskip}
\usepackage{xcolor,colortbl}
\usepackage{subcaption}
\usepackage{authblk}
\usepackage{algorithm}     
\usepackage{algpseudocode} 
\usepackage{amsmath}       
\usepackage{float}   
\usepackage{multicol}      
\usepackage[frozencache,cachedir=.,outputdir=.]{minted}
\usepackage{enumitem}
\setlist[itemize]{leftmargin=2em}
\usepackage{tikz}
\usetikzlibrary{matrix,shapes,decorations.pathreplacing,fit,backgrounds}

%% file: utils/math.tex
\usepackage{amsmath, amssymb, amsfonts, mathtools}

\usepackage{physics}

\usepackage{cancel}

\usepackage{amsthm}
\usepackage{thmtools, thm-restate}

\usepackage{accents}

\usepackage{bm}

\newcommand{\RR}{\mathbb{R}}

\newcommand{\tH}{\mathtt{D}}
\newcommand{\tL}{\mathtt{L}}
\newcommand{\tN}{\mathtt{N}}

\newcommand{\tK}{\mathtt{\ell_h}}
\newcommand{\tOne}{\mathtt{1}}
\newcommand{\tNcp}{\mathtt{N_{cp}}}
\newcommand{\tLoverNcp}{\mathtt{L // N_{cp}}}
\newcommand{\tHoverNcp}{\mathtt{H // N_{cp}}}

\renewcommand{\vec}[1]{\boldsymbol{\mathrm{#1}}}

%% file: sections/1_introduction.tex
\begin{figure}[H]
    \centering
    \includegraphics[width=0.99\textwidth]{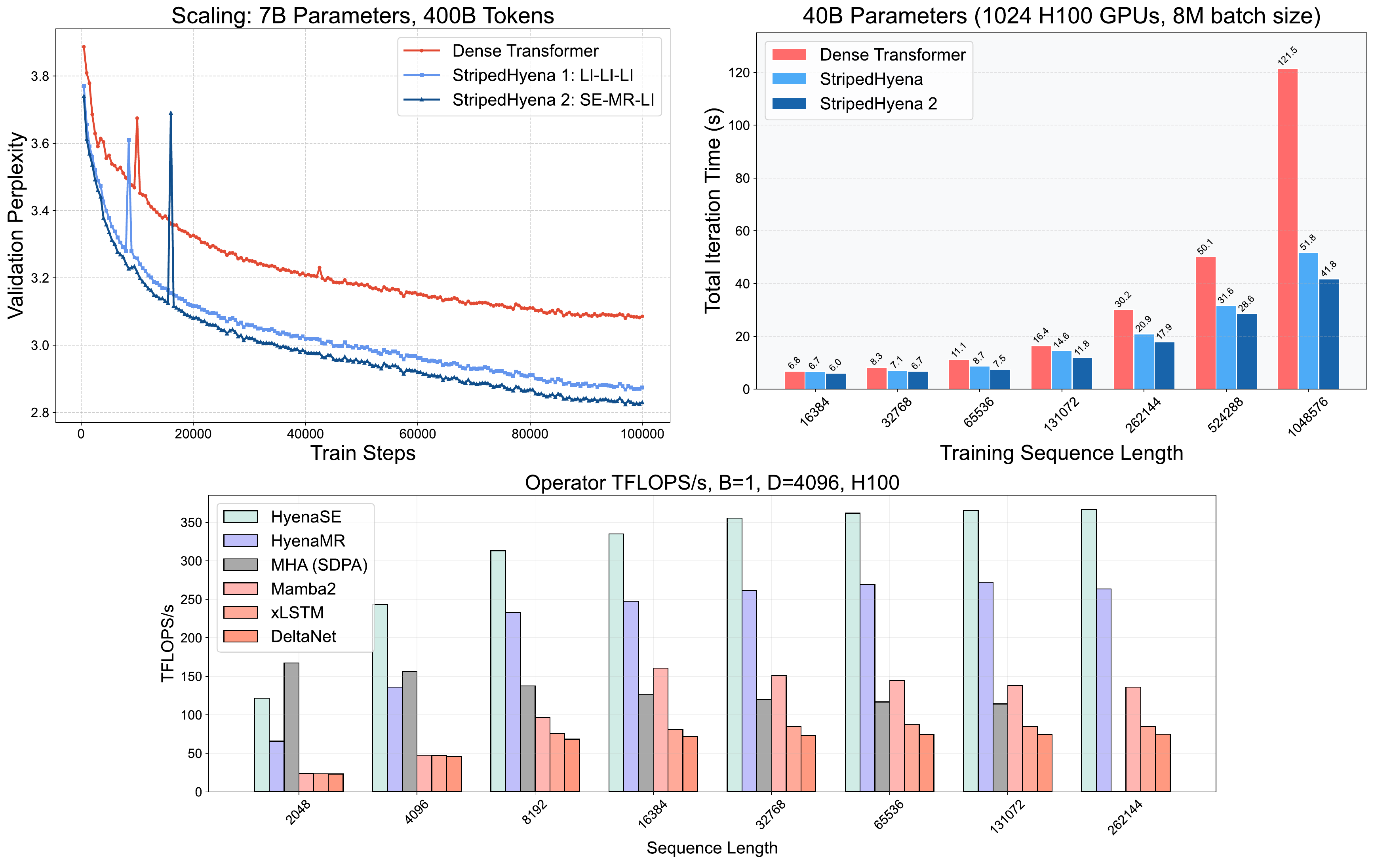}
    \vspace{-3mm}
    \caption*{Figure 1: Scaling experiments, showing differences in perplexity and throughput of Transformers, multi-hybrids (StripedHyena 2), and other alternative operators.}
    \label{fig:first}
\end{figure}

\section{Introduction}

Architecture improvements for training language models at scale can be broadly categorized into several main groups. Tweaks to the attention mechanism to reduce the size of the kv cache such as GQA, MQA, MLA, sliding window and linear attention \citep{shazeer2019fast,brown2020language,vaswani2021scaling,katharopoulos2020transformers,ainslie2023gqa,liu2024deepseek}. Changes to the model for numerical stability, resilience to outliers and quantization such as pre-norm, SwiGLU, QK normalization \citep{zhang2019root,xiong2020layer,shazeer2020glu}. Finally, modifications that improve model capacity or recall at longer context such as RoPE, MoE \citep{shazeer2017outrageously,su2024roformer}. Despite the broad interest in architecture improvement, remarkably few proposals, outside of the aforementioned methods, have delivered consistent gains at scale.

A different approach is to introduce new \textit{classes} of input-dependent operators to the standard mix of layers (self-attention and feed-forward) layers and optimize their composition, resulting in \textit{hybrid} architectures. Hybrids promise improvements on both quality and efficiency, and have been proposed in various domains and with various mixtures of operators, typically with some combination of convolution and attention \citep{dai2021coatnet,polistripedhyena}, linear attention and attention \citep{fu2022hungry,fathi2023block,lieber2024jamba,glorioso2024zamba}, or local attention and attention \citep{child2019generating,beltagy2020longformer}. In language modeling at scale, hybrids of convolutions, linear attention and attention have been validated through dedicated scaling laws \citep{poli2024mechanistic} and large-scale model releases \citep{glorioso2024zamba,nguyen2024sequence,team2024jamba}.

Despite being a promising alternative, hybrids based on operators such as linear attention or state-space models have struggled to replace baseline Transformers as the de facto standard for language modeling due to a variety of reasons. One limitation is that these fixed-state operators realize efficiency gains only when applied to very long sequences, which is also where they drastically underperform full self-attention \citep{arora2023zoology,jelassi2024repeat} in quality. Compared to Transformers, these methods are generally slower in common pretraining regimes: shorter contexts with larger and wider models. Furthermore, most of these approaches have been developed with the explicit goal of matching self-attention performance on in-context recall over longer sequences, but require hybridization with self-attention to perform competitively in practice. This has introduced redundancy in architectures, as multiple operators are optimized for the same capability: in-context recall.

This paper explores a fundamentally different approach. We advocate for model architecture designs that are both hybridization-aware and hardware-aware, combining different types of operators with complementary capabilities and computational costs, across a range of input and model sizes. Our approach is motivated by work on synthetics \citep{akyurek2024context} and mechanistic design \citep{poli2024mechanistic}, showing how different operators in hybrids can specialize to subtasks such as recall, compression, multi-query recall, and fuzzy recall. For example, input-dependent convolutions excel at filtering noise and performing multi-token recall, useful for modeling byte-level data, whereas attention is optimized for targeted recall of information across longer sequences. We introduce \textit{multi-hybrids}, architectures that combine strengths of multiple operator types. 

We focus on StripedHyena 2, the first example of a convolutional multi-hybrid architecture for sequence modeling validated on a series of experiments at scale (40 billion parameters, 9 trillion tokens). StripedHyena 2 is based on three different types of input-dependent convolutional operators: (i.) short, explicitly-parametrized hyena operators that maximize hardware utilization, specializing in local multi-token recall, (ii.) medium-length, regularized hyena operators tailored to efficient modeling across hundreds of tokens, and (iii.) long, implicit hyena operators that aggregate information over the entire sequence. We describe the algorithmic foundations of convolutional multi-hybrids, focusing on architecture design, kernels, and context parallelism algorithms. As a motivating example, we will use the experiments behind the Evo 2 line of models \citep{brixievo2}, built on top of StripedHyena 2. Evo 2 40B is a state-of-the-art foundation model for genomics, trained on byte-tokenized (nucleotide) sequences.

\paragraph{Outline}

In Section \ref{sec:architecture}, we introduce the basic design ideas and describe the three primary operators behind convolutional multi-hybrids. We then discuss composition, filter grouping for improved hardware utilization, and showcase scaling at the thousand GPU and 40 billion parameter scale. In Section \ref{sec:kernel_optimizations}, we focus on architecture and algorithm co-design. Using filter grouping, we adapt overlap-add algorithms \citep{burrus1985convolution} to tensor cores, introducing our implementation of a two-stage blocked kernel. We measure the performance gains at short and long context compared to efficient attention implementations (FlashAttention3 \citep{shah2024flashattention}, SDPA) and other alternative operators such as linear attention and state-space models: Mamba2 \citep{dao2024transformers}, xLSTM \citep{beck2024xlstm} and DeltaNet \citep{yang2024parallelizing}. In Section \ref{sec:context_parallelism}, we develop custom context parallelism methods for the different types of convolutions in our models. We introduce both peer-to-peer and all-to-all algorithms, including new channel-pipelined variants and FFT-based methods.

%% file: sections/architecture.tex
\section{Multi-Hybrid Model Architecture}\label{sec:architecture}

\paragraph{Notation}
Unless specified otherwise, (input) sequences are denoted with $x \in \RR^{\ell \times d}$. We use subscripts to index in the time dimension, and Greek superscripts to index in the space (or width) dimension. To keep the notation compact, we also occasionally omit summation signs for repeated indices in longer tensor contractions e.g., $y^\alpha_t = A^{\alpha\beta}x^{\beta}_t$ is shorthand for $y^\alpha_t = \sum_{\beta} A^{\alpha\beta}x^{\beta}_t$.

\subsection{Basic Design}

\begin{figure}[t]
    \centering
    \includegraphics[width=0.8\textwidth]{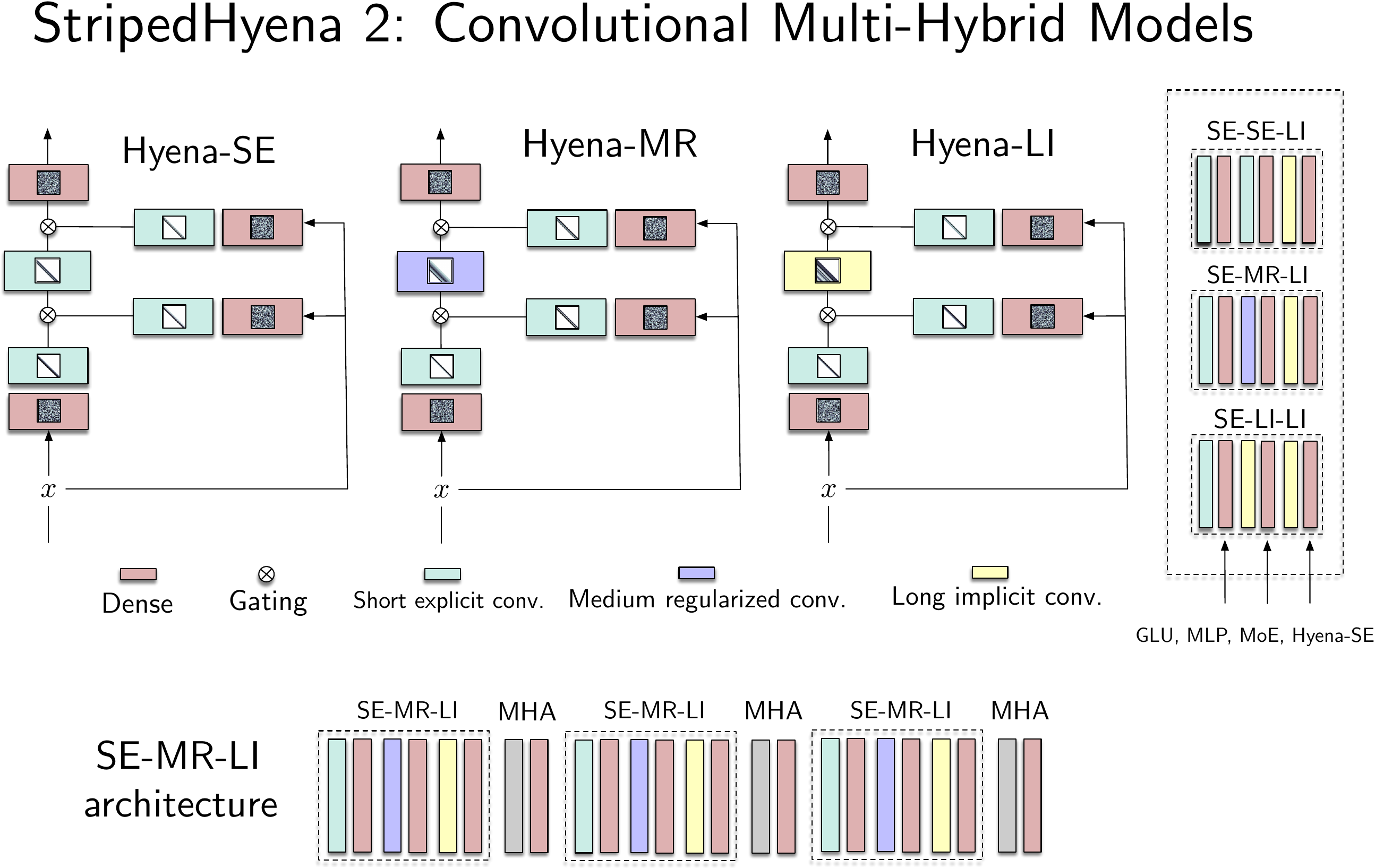}
    \caption{
    Overview of the convolutional operators forming the basis of StripedHyena 2: \textsf{Hyena-SE} (short explicit filters), \textsf{Hyena-MR} (medium regularized filters), \textsf{Hyena-LI} (long implicit filters). All operators use the Hyena structure \citep{poli2023hyena}, tailoring the inner convolution parametrization for an improved balance of quality and efficiency. Given these operators, we explore different striped layouts.}
    \label{fig:overview}
\end{figure}

We consider input-dependent convolutional operators that adhere to the following structure from the original Hyena work \citep{poli2023hyena}:

\begin{equation}\label{eq:hyena_structure}
\begin{aligned}
q_t^{\alpha} &= T^{\alpha}_{t t'} (x_{t'}^{\beta} W^{\beta \alpha}) \\ 
k_t^{\alpha} &= H^{\alpha}_{t t'} (x_{t'}^{\beta} U^{\beta \alpha}) \\ 
v_t^{\alpha} &= K^{\alpha}_{t t'} (x_{t'}^{\beta} P^{\beta \alpha}) \\
y_t^{\alpha} &= (q_t^{\beta} G^{\beta}_{t t'} k_{t'}^{\beta} v_{t'}^{\beta}) M^{\beta \alpha}
\end{aligned}
\end{equation}
where $T, H, K, G\in\RR^{d \times \ell \times \ell}$ are Toeplitz matrices (corresponding to the convolution with the respective filters $h_T, h_H, h_K, h_G$), and $W, U, P, M\in\RR^{d \times d}$ are dense matrices (parametrized as dense matrices or low-rank matrices). A schematic representation is provided in Figure \ref{fig:overview}.

In Hyena, the filters $h_T, h_H, h_K$ are parametrized explicitly: the entries of the filters are learnable parameters, analogous to the approach of classical convolutional neural networks\footnote{The presence of short explicit filters in the featurization step for query, key and value, first proposed for input-dependent convolution in \citep{poli2023hyena} and linear attention in \citep{peng2023rwkv}, has also been later adopted by other modern operator variants}. The inner filter $h_G$ is instead parametrized implicitly, with the values obtained as a combination of basis functions or as outputs of a neural network \citep{romero2021ckconv}. For this reason, computational primitives following the structure in Equation \ref{eq:hyena_structure} have been also broadly referred to as long convolution operators.

We build on this basic structure, leaning into the design of convolution operators. The main insights are that not every input-dependent convolution in a hybrid should rely on long, implicit filters, and that convolutional operators should be tailored to run fast on target hardware.

\paragraph{Input-dependent convolutional operators} The first class of input-dependent convolutions is \textsf{Hyena-LI} (long implicit), the closest relative to the original design. In \textsf{Hyena-LI}, the filters $h_T, h_H, h_K$ remain short and explicit, while the inner filter is obtained as a linear combination of real exponentials $h_t = \sum_{n=1}^{d} R_n \lambda_n^{t-1}, R_n, \lambda_n \in \RR$ \citep{massaroli2024laughing}. This is a real-valued, simplified version of a variety of other parametrizations \citep{orvieto2023resurrecting,gupta2022diagonal}, with the addition of filter grouping (Section \ref{sec:additional_design_decisions}). Due to this choice, \textsf{Hyena-LI} retains the ability to switch to an recurrent parametrization for constant memory.

Next, we define \textsf{Hyena-SE} (short explicit), a variant with short, explicit filters in all its convolutions. When the filters are short\footnote{In our experiments, shorter than 14. In our final runs at scale, we used a range of 4 to 7.}, a simple explicit parametrization is sufficient to achieve convergence. \textsf{Hyena-SE} is key in achieving speedups across a range of input regimes, including short sequences, while still excelling at local, multi-token recall. With a hardware-aware implementation using tensor cores, \textsf{Hyena-SE} achieves the highest throughput of any sequence mixing operator (Section \ref{sec:kernel_optimizations}). \textsf{Hyena-SE} can also be utilized as a replacement for feed-forward layers.

Finally, we introduce \textsf{Hyena-MR} (medium regularized), a variant with explicitly parametrized filters of length in the hundreds. While it can be difficult to optimize longer explicit convolutions, we find that a simple exponential decay regularizer i.e., $h_t = \hat{h}_t \lambda^{-\alpha t}$, where $\alpha$ is swept across channels and $\hat{h}_t$ is the learnable parameter, is sufficient for convergence. With filter grouping and an efficient implementation using tensor cores, this variant remains significantly faster than linear attention and state-space models (Section \ref{sec:kernel_optimizations}). \textsf{Hyena-MR} is to \textsf{Hyena-LI} what sliding window attention is to the classic attention operator.

Since the filters in \textsf{Hyena-MR} and \textsf{Hyena-SE} are finite-impulse response (FIR), these operators trivially retain constant memory during autoregressive generation, analogous to sliding window attention.

Multi-hybrids interleave \textsf{Hyena-SE}, \textsf{Hyena-MR} and \textsf{Hyena-LI} to obtain the full architecture.









\subsection{Additional Design Decisions}\label{sec:additional_design_decisions}

Next, we detail additional design aspects of convolutional multi-hybrids, including block layout, weight-sharing filter patterns, and effectiveness of context extension techniques. 

Experiment configuration files and code are provided in \textsf{Savanna}, our fully open-source pretraining infrastructure for research on multi-hybrids: \href{https://github.com/Zymrael/savanna}{{\tt https://github.com/Zymrael/savanna}}.

\paragraph{Multi-hybrid block layout}
%


We explore the effect of block layouts on performance and training speed. Table \ref{tab:block_layouts} shows the results of training 7B multi-hybrid models on 400B tokens of {\tt OpenGenome2} \citep{brixievo2}, each with different block layouts. We use our default inner filter lengths of 7 for \textsf{SE} and 128 for \textsf{MR}.

\begin{wraptable}[11]{r}{0.4\columnwidth}
    \centering
    \begin{tabular}{lcccc}
        \toprule
        \rowcolor{blue!10}\textbf{\textsf{Layout}} & \textbf{\textsf{PPL@400B}} \\
        \midrule
        \textsf{MHA-MHA-MHA} & 3.09 \\
        \midrule
        \textsf{LI-LI-LI} & 2.87 \\
        \textsf{SE-SE-LI} & 2.88 \\
        \textsf{SE-MR-LI} & \underline{2.83} \\
        \bottomrule
    \end{tabular}
    \caption{Effect of different block layouts on pretraining at the 7B parameter scale.}
    \label{tab:block_layouts}
\end{wraptable}

The blocks are repeated until target 7B parameter model depth (32) is achieved. All StripedHyena 2 models in addition interleave 5 MHA operators with the convolutional blocks. Validation perplexity measured after training of 400B tokens of byte-tokenized data (DNA sequences from {\tt OpenGenome2}), with 7B parameter StripedHyena 2 models.

At this scale, \textsf{SE-MR-LI} perform best on pretraining quality. Notably, we find that pure long convolution \textsf{LI-LI-LI} layouts can be replaced by \textsf{SE-SE-LI} blocks for little-to-no loss in quality and significant benefits to throughput. While \textsf{SE-MR-LI} block layouts provide a general stable baseline for multi-hybrids, we recommend ablating block layouts for new tasks or domains, particularly if parametrization hyperparameters such as filter length, decay strength, and initialization are modified.
%

%

\paragraph{Weight-sharing filter patterns}
We propose a grouped\footnote{Note that this approach is not the same as traditional grouped CNN layers, which instead mixes across channels in the same group.} design for input-dependent convolutional operators, where the filters are shared across groups of channels. Namely, let $\mathcal{G}$ be a group of channels of size $d_g$. Then,
\[
\forall \alpha \in \mathcal{G}: y^\alpha_t = \sum_{j=0}^{t} h_{t-j}^\mathcal{G} x_{j}^\alpha.
\]
The main benefit of our grouping is to enable efficient representation of the discrete convolution as a series of general matrix-matrix multiplications ({\tt GEMM}) instead of general matrix-vector multiplications ({\tt GEMV}). We use this property to co-design hardware-aware algorithms for our architecture (see Section \ref{sec:kernel_optimizations}). Grouping has minimal effect on quality (Section \ref{sup-fig:overlapping_comm}).




%


\paragraph{Context extension}

We tested context extension up to 1 million sequence length on 7B and 40B multi-hybrids using techniques developed for rotary attention, such as \textit{position interpolation} PI \citep{chen2023extending} and \textit{adjusted base frequency} (ABF) \citep{xiong2023effective} and a combination of both. We found only minor differences in perplexity (Table \ref{tab:context_extension}), with all models capable of in-context recall at the target maximum context length. Recall results are shown in Figure \ref{fig:niah_extension}.

\begin{table}[H]
    \centering
    \begin{tabular}{lcccccc}
        \toprule
        \rowcolor{blue!10}\textbf{\textsf{Extension Method}} & \multicolumn{6}{c}{\textbf{\textsf{Context Length (K)}}} \\ \cmidrule{2-7}
        & \textbf{32} & \textbf{65} & \textbf{131} & \textbf{262} & \textbf{524} & \textbf{1048} \\ 
        \midrule
        \textsf{Position Interpolation (PI)} & 2.785 & 2.763 & 2.750 & - & - & - \\
        \textsf{PI + ABF} & 2.782 & 2.763 & 2.748 & 2.707 & 2.663 & 2.597 \\
        \bottomrule
    \end{tabular}
    \caption{Validation perplexity of 7B StripedHyena 2 architecture on {\tt OpenGenome2} after midtraining extension with different techniques, terminating in extension at 1 million context length. Midtraining is performed on the base StripedHyena 2 7B trained on 2T tokens of {\tt OpenGenome2} at byte resolution and 8192 context length (Evo 2 7B). The values are collected at the end of training, for a visualization of the trends see Figure \ref{fig:niah_extension}.}
    \label{tab:context_extension}
\end{table}


\subsection{Measuring Throughput at Scale}

Owing to its design incorporating FIR convolution operators such as \textsf{Hyena-SE} and \textsf{Hyena-MR}, multi-hybrids achieve consistent speedups compared to previous generation hybrids and Transformers.

\begin{figure}[H]
    \includegraphics[width=0.99\textwidth]{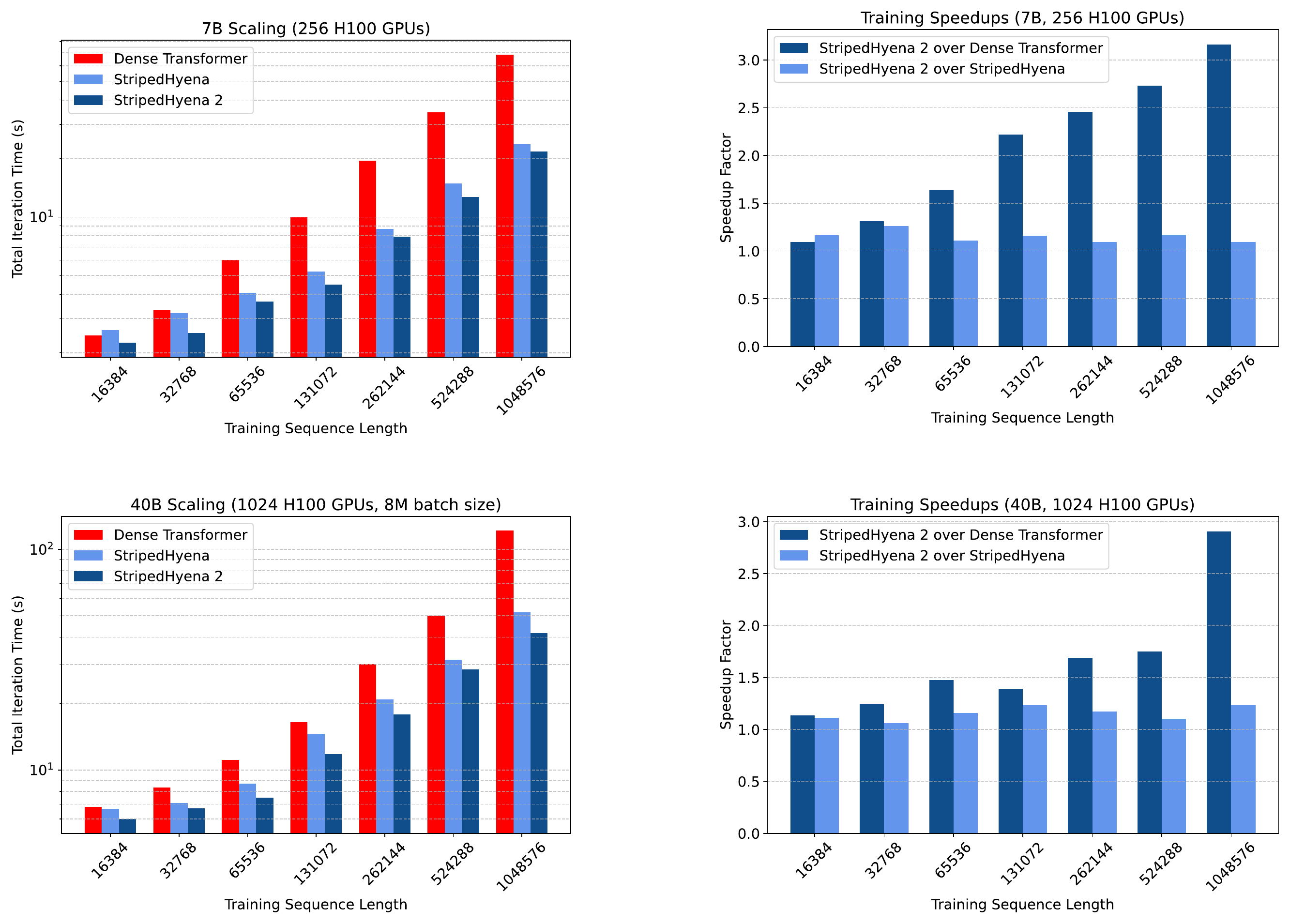}
    \vspace{-0.1cm}
    \caption{End-to-end iteration times (forward and backward) during training, collected on a large cluster of H100 SXM GPUs. See Table \ref{tab:appendix_scaling} for details on the measurement protocol.}
    \label{fig:all_scaling}
\end{figure}

Across the 7B and 40B parameter scales, StripedHyena 2 trains 1.2 to 2.9 times faster on a H100 cluster compared to our optimized Transformer based on a reference {\tt Transformer Engine} implementation, collected during training with FP8 precision on dense (SwiGLUs, projections) and normalization layers (Figure \ref{fig:all_scaling}). Notably, it also achieves speedups at shorter sequence lengths, compared to both Transformers and StripedHyena \citep{polistripedhyena}, a previous generation hybrid. Given an comparable degree of optimization in the implementations, we expect similar or better gains on other training infrastructure. See Table \ref{tab:appendix_scaling} for details on the measurement protocol.




We also report TFLOPS per second per GPU and MFU\footnote{We use a reference number of 1000 TFLOPs per H100.} in Figure \ref{fig:all_scaling}. Since we use the same distributed settings for all architectures, we note a lower MFU for hybrids at longer sequence lengths. This is primarily due to a reduction in overall model FLOPS\footnote{We used actual model FLOPS instead of approximations, since most approximations are not accurate at very long context. For self-attention FLOPS, we used the estimate in \cite{dao2023flashattention}.} caused by subquadratic scaling in sequence length. Consequently, we expect further tuning of distributed settings for multi-hybrids at long context to yield even larger speedups by using e.g., larger micro batch sizes to increase compute intensity of each rank. Further discussion on context parallelism implementation is provided in Section \ref{sec:context_parallelism}.

Convolutional multi-hybrids such as StripedHyena 2 similarly outscale other previous-generation hybrids, including those based on linear attention or state-space models. Latency and MFU at the operator level are provided in the following section.




%% file: sections/kernel_optimizations.tex
\section{Hardware-Aware Convolution Algorithms}\label{sec:kernel_optimizations}
Discrete convolutions are mathematically equivalent to matrix multiplications with Toeplitz matrices\footnote{Algorithms for fast convolutions correspond to fast matrix multiplications schemes for Toeplitz matrices.}. Convolutional operators in StripedHyena 2 operate in computational regimes that are quite different from traditional convolutional neural networks (CNNs), requiring custom algorithms:
\begin{itemize}
    \item {\tt Hyena-SE}: \textit{short-explicit} hyena layers are based on a combination of grouped depthwise convolutions with explicitly parametrized shorter filters (e.g., length 7).
    \item {\tt Hyena-MR}: \textit{medium-regularized} hyena layers are based on a combination of grouped depthwise convolutions with longer filters (e.g., length 128), obtained by applying a regularization term to the filter weights.
    \item {\tt Hyena-LI}: \textit{long-implicit} hyena layers are based on a combination of grouped depthwise convolutions with longer \textit{implicit} filters (as long as the sequence length) \citep{romero2021ckconv,poli2023hyena}, in the same family as other long convolutions.
\end{itemize} 
Libraries such as PyTorch provide optimized implementations of short explicit convolutions via a variety of backends due to their common utilization in CNN architectures. These implementations span approaches such as im2col and Winograd algorithms \citep{chetlur2014cudnn,vasudevan2017parallel}. However, existing implementations are not fully optimized for the convolution operators used in multi-hybrids, which rely mostly on depthwise convolutions of different lengths. When the convolution filter is very long, most algorithms rely on FFT-based methods. These approaches are known to suffer from lower hardware utilization, despite modifications to leverage tensor cores \citep{li2021tcfft,fu2023flashfftconv}. Instead of im2col or Winograd GEMM algorithms for convolutions, we focus on a direct multi-pass blocked approach that is co-designed to exploit filter grouping in Hyena.


%

\subsection{Block Convolution}
\label{sec:block_conv_background}
For a causal FIR filter of length $\ell_h$ applied to an input signal $x$ of length $\ell$ (with typical $\ell \gg \ell_h$), the output at index $t$ is
\begin{equation}\label{eq:fir_def}
    y_t = \sum_{k=0}^{t} h_{t-k}\,x_k, \quad \text{where } h_k=0 \text{ for } k<0 \text{ or } k\ge \ell_h.
\end{equation}
This can be written in matrix form:
\begin{equation}\label{eq:toeplitz_matrix}
    \begin{bmatrix}
        y_0 \\
        y_1 \\ 
        y_2 \\
        \vdots \\
        y_{\ell-1}
    \end{bmatrix}
    =
    \underbrace{
        \begin{bmatrix}
            h_0   & 0     & 0     & 0     & 0     & \cdots & 0     \\
            h_1   & h_0   & 0     & 0     & 0     & \cdots & 0     \\
            \vdots& h_1   & h_0   & 0     & 0     & \cdots & 0     \\
            h_{\ell_h-1} & \vdots & h_1   & h_0   & 0     & \cdots & 0     \\
            0     & h_{\ell_h-1} & \vdots & h_1   & h_0   & \cdots & 0     \\
            \vdots     & \ddots & \ddots & \ddots & \ddots & \ddots & 0 \\
            0     & \cdots & 0     & h_{\ell_h-1} & \cdots  & h_1 & h_0   
        \end{bmatrix}
    }_{\displaystyle T}
    \begin{bmatrix}
        x_0 \\
        x_1 \\
        x_2 \\
        \vdots \\
        x_{\ell-1}
    \end{bmatrix}.
\end{equation}
Classical digital signal processing often handles FIR filters using \emph{block convolution} methods \citep{burrus1985convolution}. One partitions both the input and the output signals into chunks of size $\ell_b$, then multiplies $\ell_b \times \ell_b$ sub-blocks of $T$ against these smaller segments. Once the filter's support is exceeded, many sub-blocks are purely zeros and can be skipped -- an advantage when $\ell_h \ll \ell$.

Concretely, the input and output sequences are chunken into $x=(\hat x_0, \hat x_1, \dots)$, $y=(\hat y_0, \hat y_1, \dots)$ with  
\begin{equation}
    \hat x_k = (x_{k\ell_b},  \dots, x_{k\ell_b+\ell_b-1}), \quad 
    \hat y_k = (y_{k\ell_b}, \dots, y_{k\ell_b+\ell_b-1}), \quad k = 0, 1, 2, \dots, \lceil \ell/\ell_b \rceil-1.
\end{equation}
The fully partitioned Toeplitz matrix factors into submatrices $H_0, H_1, \dots, H_{\lceil \ell/\ell_b \rceil-1}$ of size $\ell_b \times \ell_b$. For instance,
\begin{equation}
    H_0 = \begin{bmatrix}
        h_0 & 0 & \cdots & 0 \\
        h_1 & h_0 & \ddots & \vdots \\
        \vdots & \ddots & \ddots & 0 \\
        h_{\ell_b-1} & \cdots & h_1 & h_0
    \end{bmatrix}, 
    \quad
    H_1 = \begin{bmatrix}
        h_{\ell_b} & h_{\ell_b-1} & \cdots & h_1 \\
        h_{\ell_b+1} & h_{\ell_b} & \ddots & \vdots \\
        \vdots & \ddots & \ddots & h_{\ell_b-1} \\
        h_{2\ell_b-1} & \cdots & h_{\ell_b+1} & h_{\ell_b}
    \end{bmatrix}, \quad \dots
\end{equation}
Hence,
\begin{equation}
    T = \begin{bmatrix}
        H_0      &        &        & \\
        H_1      & H_0    &        & \\
        H_2      & H_1    & H_0    & \\
        \vdots   & \vdots & \vdots & \ddots \\
        
    \end{bmatrix}.
\end{equation}
Since $h_t=0$ for $t\ge \ell_h$, any blocks with index greater than $\left\lceil (\ell_h-1)/\ell_b \right\rceil+1$ yields a zero submatrix. Hence, we only need to construct and multiply the non-zero submatrices $H_k, k=0, 1, \dots, \left\lceil (\ell_h-1)/\ell_b \right\rceil$.
The output blocks then obey:
\begin{equation}\label{eq:block_conv_def}
    \hat y_n = \sum_{k=0}^{n} H_{n-k}\,\hat x_k = \sum_{k=0}^{\lceil (\ell_h-1)/\ell_b \rceil} H_{k}\,\hat x_{n-k}.
\end{equation}

Note that \emph{block convolution} \eqref{eq:block_conv_def} can be seen as a “convolution of convolutions”, since each block $H_k$ is itself a Toeplitz matrix and can be implemented efficiently by direct multiplication or by using fast convolution techniques (e.g., FFT-based methods when $\ell_b$ is large) within each block. 

\subsection{Simple Two-Stage Block Algorithm}
\label{sec:two_stage_algo}

For many \texttt{Hyena-SE} or \texttt{Hyena-MR} use cases, $\ell_h$ (the filter length) is much smaller than $\ell$ (the sequence length). When $\ell_h$ is also within about twice the chosen block size $\ell_b$, a particularly efficient two-stage block algorithm can be used. 

Let $T$ be the Toeplitz matrix that applies a grouped depthwise FIR filter of length $\ell_h$. Suppose we choose a block size $\ell_b$ such that $\ell_h \le 2 \ell_b$. Under this condition, $T$ can be decomposed into a block-diagonal part plus an off-diagonal part (or ``stage''):
\begin{equation}
    T = \underbrace{\begin{bmatrix}
       H_0 &  &  & \\
         & H_0 &  & \\
         &  & \ddots &  \\
         &  &  & H_0
    \end{bmatrix}}_{\text{first stage}}
    \;+\;
    \underbrace{\begin{bmatrix}
         &  &  & \\
        H_1 &  &  & \\
         &  \ddots&  &  \\
         &  &  H_1 & 
    \end{bmatrix}}_{\text{second stage}}.
\end{equation}
In particular:
\begin{itemize}
    \item $H_0$ covers the points of the filter that align with the current chunk $\hat x_k$.
    \item $H_1$ covers the points of the filter that spills over from the previous chunk $\hat x_{k-1}$, capturing taps that straddle the boundary between adjacent chunks.
\end{itemize}

\begin{note}{colback=teal!10}
As an illustrative example, consider $\ell=6$ (sequence length), $\ell_h=4$ (filter length), and $\ell_b=3$ (block size). The filter coefficients $h_0, h_1, h_2, h_3$ form the blocks:
\[
H_0 = \begin{bmatrix}
h_0 & 0 & 0 \\
h_1 & h_0 & 0 \\
h_2 & h_1 & h_0
\end{bmatrix}, \quad
H_1 = \begin{bmatrix}
h_3 & h_2 & h_1 \\
0 & h_3 & h_2 \\
0 & 0 & h_3
\end{bmatrix}.
\]
The full Toeplitz matrix $T$ decomposes as:
\[
T = \underbrace{\begin{bmatrix}
H_0 & 0 \\
0 & H_0
\end{bmatrix}}_{\text{first stage}}
+
\underbrace{\begin{bmatrix}
0 & 0 \\
H_1 & 0
\end{bmatrix}}_{\text{second stage}}
= \begin{bmatrix}
h_0 & 0 & 0 & 0 & 0 & 0 \\
h_1 & h_0 & 0 & 0 & 0 & 0 \\
h_2 & h_1 & h_0 & 0 & 0 & 0 \\
h_3 & h_2 & h_1 & h_0 & 0 & 0 \\
0 & h_3 & h_2 & h_1 & h_0 & 0 \\
0 & 0 & h_3 & h_2 & h_1 & h_0
\end{bmatrix}.
\]
\end{note}

This approach presents a number of advantages.  Once loaded, $H_0$ and $H_1$ can be reused across multiple chunks of the input and, with our grouped operator design, also across multiple channels within the same group. This provides a convenient way to turn small {\tt GEMV} operations into {\tt GEMMs}, compared to other {\tt GEMM} approaches for convolutions that rely on forming strided views of the input. Furthermore, the decomposition separates “current chunk” vs. “previous chunk” computations, which can be run in parallel or as a pipeline. 


\paragraph{Analysis of the two-stage multiplication.}
Suppose we denote by $\hat{X}_n \in \RR^{\ell_b \times d}$ the $n$-th input chunk (as in \eqref{eq:block_conv_def}), where $d$ denotes both the group size and the tensor core dimension, and let $\hat{Y}_n \in \RR^{\ell_b \times d}$ be the corresponding output chunk. Under a two-stage block convolution, each $\hat{Y}_n$ is computed as
\begin{equation}\label{eq:two_stage_formula}
    \hat{Y}_n = H_0\,\hat{X}_n \;+\; H_1\,\hat{X}_{n-1},
    \quad
    n = 0, 1, \dots, \lceil \ell/\ell_b \rceil - 1,
    \quad
    \text{(with $\hat{X}_{-1} = 0$ for $n=0$).}
\end{equation}
Here, $H_0$ captures the filter taps interacting with the “current” chunk $\hat{X}_n$, while $H_1$ handles the “spillover” from the preceding chunk $\hat{X}_{n-1}$. By construction, $\ell_h \le 2\,\ell_b$ ensures that no additional off-diagonal blocks appear beyond $H_1$.

In the setting where all channels within a group share the same filter, the matrices $H_0 \in \RR^{\ell_b \times \ell_b}$ and $H_1 \in \RR^{\ell_b \times \ell_b}$ are common to every channel in the group. Consequently, instead of forming a block-diagonal structure over separate channels, one can directly operate on the full input block $\hat{X}_n \in \RR^{\ell_b \times d_g}$, where $d_g$ is the group size. In particular, if the tensor core is of size $d_g$ and $\ell_b = d_g$, then \eqref{eq:two_stage_formula} can be implemented as two full {\tt GEMM} operations. This approach maximizes throughput and fully leverages tensor core utilization by processing the entire group in one efficient matrix multiplication. 

\subsubsection{Implementation}

We report the kernel implementation of our two-stage blocked algorithm in Algorithm \ref{alg:two_stage_chunked}. The crux lies in the grouping structure, which enables data reuse and efficient computation using tensor cores dedicated hardware units on NVIDIA GPUs specialized for high throughput matrix multiplication. Without grouping, one would need to adopt a different strategy to transform depthwise convolutions from {\tt GEMVs} to {\tt GEMMs} to avoid lower throughput CUDA cores, for example forming an input view that parallelizes the first-stage multiplication with $H_0$ across all blocks.



\begin{algorithm}[H]
\caption{Simple Two-Stage Blocked Hyena Convolution (Forward)}
\begin{algorithmic}[1]
\Require Input $v,~q,~k \in \RR^{\ell \times d}$, filter $h \in \RR^{\ell_h}$, block size $\ell_b$
\Ensure Output $y \in \RR^{\ell \times d_g}$ 
\State Chunk inputs $v,~q,~k$ into blocks $v_i,~q_i,~k_i$ of size $\ell_b \times d_g$
\For{block $i = 0$ to $\lceil \ell/\ell_b \rceil - 1$}
    \State Load $v_i,~q_i,~k_i,~H_0,~H_1$ to on-chip memory
    \State Initialize $y_i = 0$
    \State Optional: $v_i \gets k_i \odot v_i$
    \State $y_i \gets H_0 v_i$ \Comment{First {\tt GEMM}: block-diagonal}
    \If{$i > 0$}
        \State Load $x_{i-1}$ to on-chip memory
        \State $y_{i} \gets y_{i} + H_1 v_{i-1}$ \Comment{Second {\tt GEMM}: off-diagonal}
    \EndIf
    \State Optional: Compute $y_{i} \gets q_i \odot y_i$
\EndFor
\State \Return $y$
\end{algorithmic}
\label{alg:two_stage_chunked}
\end{algorithm}


\subsubsection{Profiling}

Convolutional multi-hybrids models are designed to be efficient across a wide range of regimes, compared to both full attention and other subquadratic operators. We optimize for both short and long sequences, as pretraining is often performed at shorter sequence lengths to maximize throughput.

\paragraph{Measurement protocol}

\begin{figure}[H]
    \centering
    \includegraphics[width=0.49\textwidth]{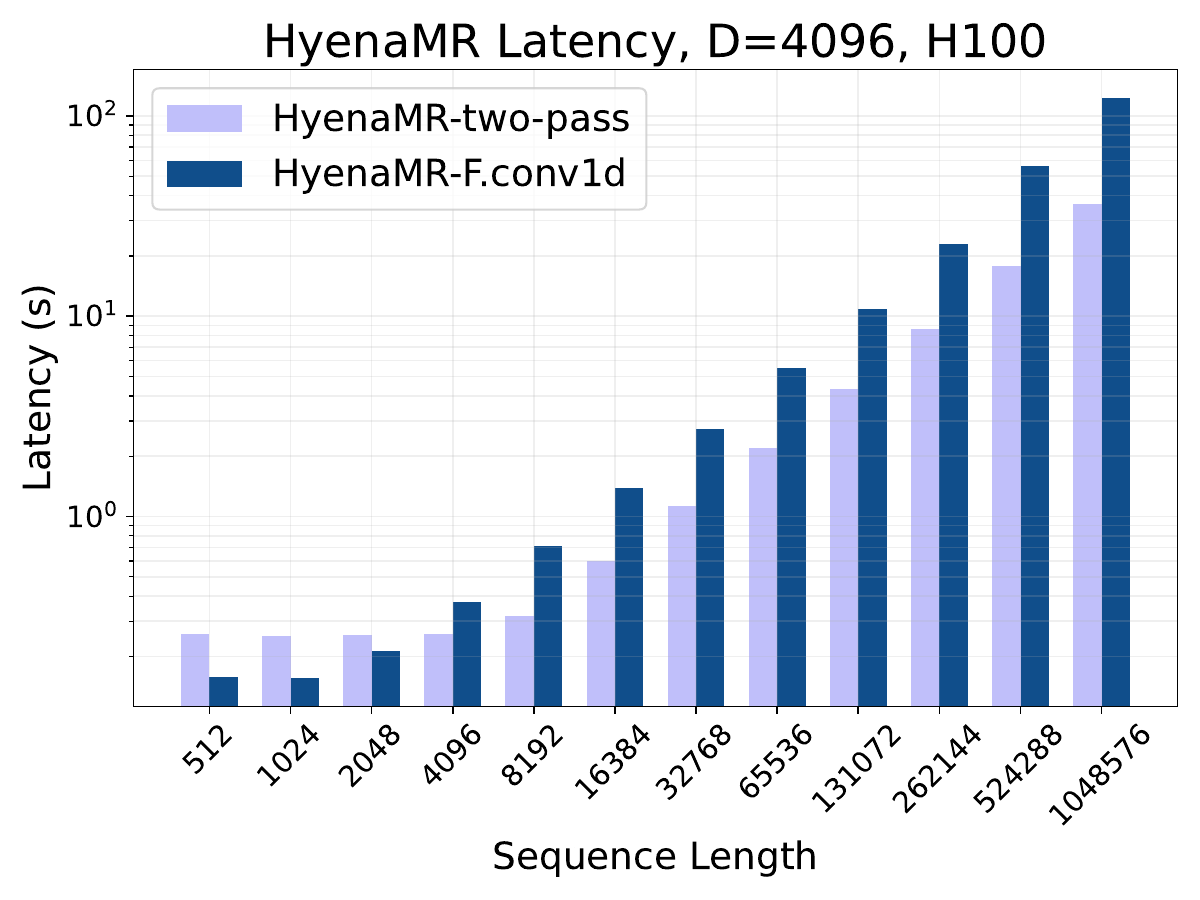}
    \includegraphics[width=0.49\textwidth]{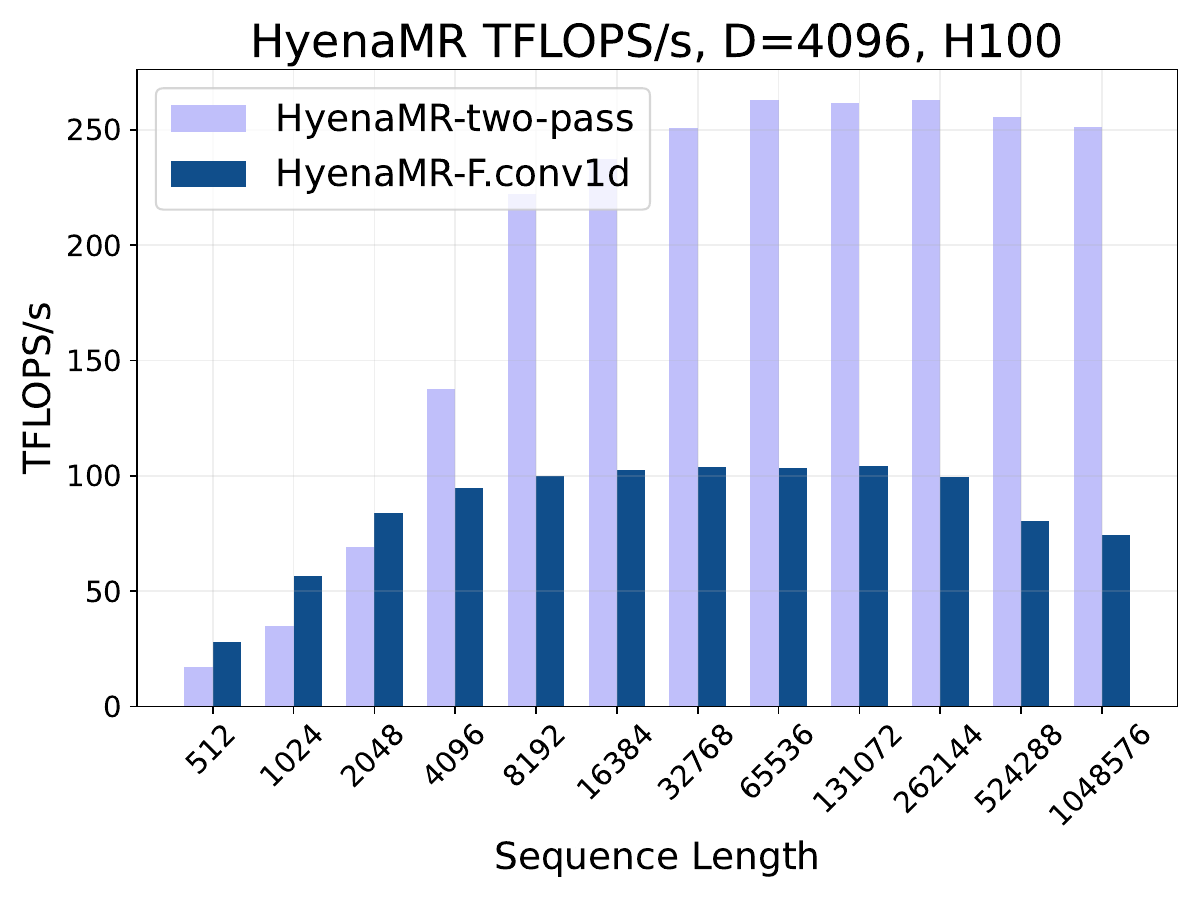}
    \vspace{-2.5mm}
    \caption{Forward latency and TFLOPS / second of \textsf{Hyena-MR} variants with filter length $128$. We compare a baseline implementation using PyTorch convolutions and our two-stage blocked kernel, showing substantial improvements in latency and throughput.}
    \label{fig:hyenamr_variants}
\end{figure}

\begin{figure}[H]
    \centering
    \includegraphics[width=0.9\textwidth]{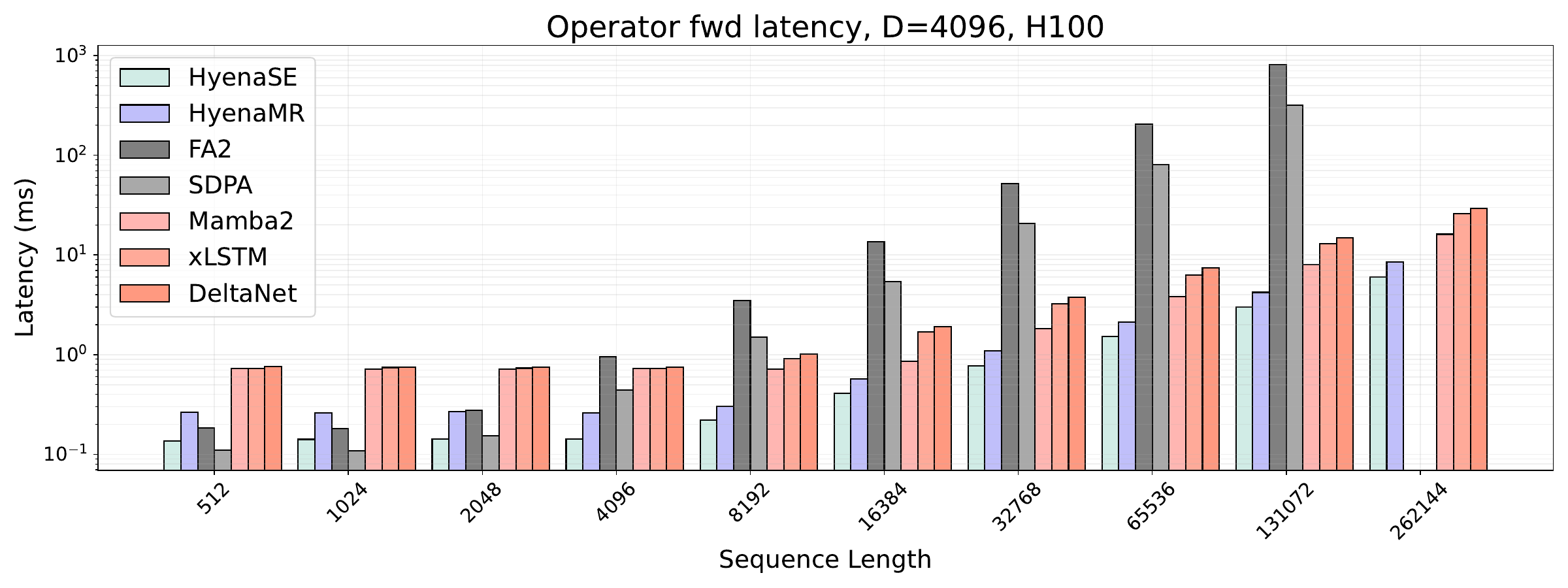} \\
    \caption{Forward latency and TFLOPs / second of \textsf{Hyena-SE}, \textsf{Hyena-MR} and other common operators in architecture design: \textit{multi-head attention} (MHA) and linear attention variants. All values are collected at operator width $4096$ (corresponding to model width at $7$B parameters), on H100s. For MHA, we report both a highly optimized implementation for Hopper GPUs (PyTorch SDPA) as well as a previous generation implementation not optimized for Hopper GPUs (FlashAttention2) \citep{dao2023flashattention}. All other operators use their official auto-tuned Triton kernels. Convolutional primitives remain efficient across sequence lengths, with substantially higher throughput than other operators, including efficient alternatives to MHA.}
    \label{fig:operator_profiles}
\end{figure}

We measure the latency and throughput of common operators in a batch size 1, varying sequence length, model width 4096 setting (corresponding to common operator width for 7B models), including input and output projections. We do not keep the total number of tokens constant, in contrast to the protocol used in FlashAttention 3 \citep{shah2024flashattention}. Figure \ref{fig:operator_profiles} and \ref{fig:tflop-app} provide the results.

%% file: sections/context_parallelism.tex
\section{Training Multi-Hybrids on Long Sequences}\label{sec:context_parallelism}

\paragraph{Notation}

In the following, we consider an input of shape $[\tOne, \tH, \tL]$, with batch size $\tOne$, hidden size $\tH$ and length $\tL$ and omit the leading $\tOne$. The discussion can be safely extended to inputs with larger batch sizes following standard data parallelism. For a CP group consisting of $\tNcp$ devices, the input is sharded along the sequence dimension and split across each of the devices in the group, so that each rank holds an input shard of shape $[\tH, \tLoverNcp]$.

\subsection{Background}
\textit{Context parallelism} (CP) refers to a collection of distributed training techniques designed to handle the growing size of models and the increasing dimension of their inputs by processing segments of the full input sequence. Context parallelism complements other distributed training techniques such as data parallelism, tensor parallelism, sequence parallelism\footnote{Context and sequence parallelisms refer to different techniques by popular convention. Sequence parallelism distributes the sequence outside tensor parallel regions e.g., normalization layers}, pipeline parallelism and other strategies for partitioning of model parameters, gradients and optimizer states \citep{rajbhandari2020zero,zhao2023pytorch}.


\paragraph{\texttt{All-to-all} context parallelism} 

In \texttt{a2a} context parallelism, each device is allowed to exchange data with every other device in the context parallel group to reconstruct the entire input sequence and hold hidden dimension splits on each CP rank instead. Concretely, the $\tNcp$ shards of shape $[\tH, \tLoverNcp]$ are redistributed among all devices, such that each device ends up with shards of shape $[\tHoverNcp \tL]$ instead. This allows each rank to independently carry out sequence mixing (e.g., attention, or convolutions). It is important to emphasize that the hidden dimension must be split in such a way that no additional communication is required to successfully complete the operation. Extended background is provided in Section \ref{sup-sec:ulysses}.


\paragraph{\texttt{Point-to-point} context parallelism} 

While \texttt{a2a} CP allows processing long inputs across multiple devices, the cost of running the operator over the whole sequence can still be very expensive. Furthermore, naive {\tt a2a} can lower utilization without appropriate overlap of communication and computation, as {\tt a2a} calls can take a significant amount of time with larger message sizes. To overcome some of these issues, \textit{point-to-point} ($\texttt{p2p}$) context parallelism allows ranks to exchange data directly with a single peer at a time rather than broadcasting to all devices. These schemes essentially perform several rounds of blocked computation and communication. 
Section \ref{sup-sec:p2p-att} describes the common ring-based algorithm for attention \citep{liu2023ring}.

\subsection{Context Parallel Hyena Operators}

Sequence mixing in Hyena operators is implemented via convolutions with different filter lengths, which need to be addressed appropriately when implementing context parallelism. We first present the general \texttt{a2a} and \texttt{p2p} CP formulations for general causal convolutions, followed by modifications specifically tailored to FFT convolutions.

\begin{figure}[H]
    \centering
    \includegraphics[width=0.99\textwidth]{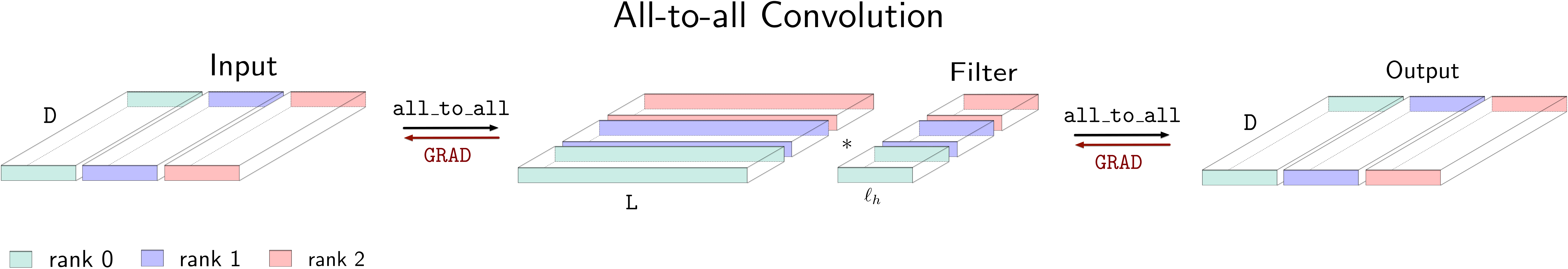}
    \caption{Diagram of computation and communication in \texttt{all-to-all convolutions}. This context parallelism strategy can be used in both inner hyena convolutions (corresponding to multiplication with $G_{t t'}$,  Eq. \ref{eq:hyena_structure}) or featurizer convolutions ($T_{tt'}$, $H_{tt'}$, $K_{tt'}$). Filters are stored or computed in each context parallel rank to avoid communication overheads. The convolution inside the context parallel region can be computed with any algorithm e.g., FFT-based or direct.}
    \label{fig:a2a_convs}
\end{figure}

\paragraph{\texttt{All-to-all} convolutions (Fig.~\ref{fig:a2a_convs})}

Let us consider an input sharded along the sequence dimension over $\tNcp$ ranks, such that each rank holds a split of shape $[\tH, \tLoverNcp]$. Analogous to the general \texttt{a2a} context parallel formulation, we perform communication across all CP ranks so that input shards of shape $[\tHoverNcp, \tL]$ are held on each device, then convolve\footnote{Both causal and non-causal convolutions are supported in this case.} each shard within the context parallel region. Finally, we perform an additional {\tt a2a} operation.

For convolutional operators in multi-hybrids, additional considerations are needed. First, filters can be stored or materialized directly inside each context parallel region. In \textsf{Hyena-SE}, each context parallel rank stores $\tHoverNcp$ filters in the depthwise case, without grouping. Care must be taken to ensure filter groups are not split across context parallel ranks. In \textsf{Hyena-MR} and \textsf{Hyena-LI}, computation of the filters can be run in each context parallel region, keeping implicit or regularization parameters sharded. If {\tt a2a} parallelism is used as the scheme of choice for the inner convolution, gating can be performed outside the context parallel region, to avoid communication overheads.


During the backward pass, gradients must be sent back to their incoming ranks during the forward call. Consequently, this involves calling two additional {\tt a2a} calls, plus the backward function on the convolutional operation on each rank.

\begin{note}{colback=blue!5}
    \textbf{[Extension] \texttt{All-to-all} channel-pipelined convolutions:} One drawback of {\tt a2a} methods is that communication latency can create bottlenecks when the message size is large, with a small fraction of time spent on compute. For {\tt a2a}, this occurs when model size and sequence length grows. One strategy is to pipeline {\tt a2a} calls and asynchronously overlap compute and communication using CUDA streams. Instead of pipelining over sequence length \citep{yao2024training}, we explore pipelining over channels to hide some of the communication latency. Concretely, we chunk inputs $[\tH, \tLoverNcp]$ into ${\tt N_{pipe}}$ segments and run an asynchronous loop of {\tt a2a} calls, scheduling to overlap compute and following {\tt a2a} call.
\end{note}

\begin{figure}[H]
    \centering
    \includegraphics[width=0.75\textwidth]{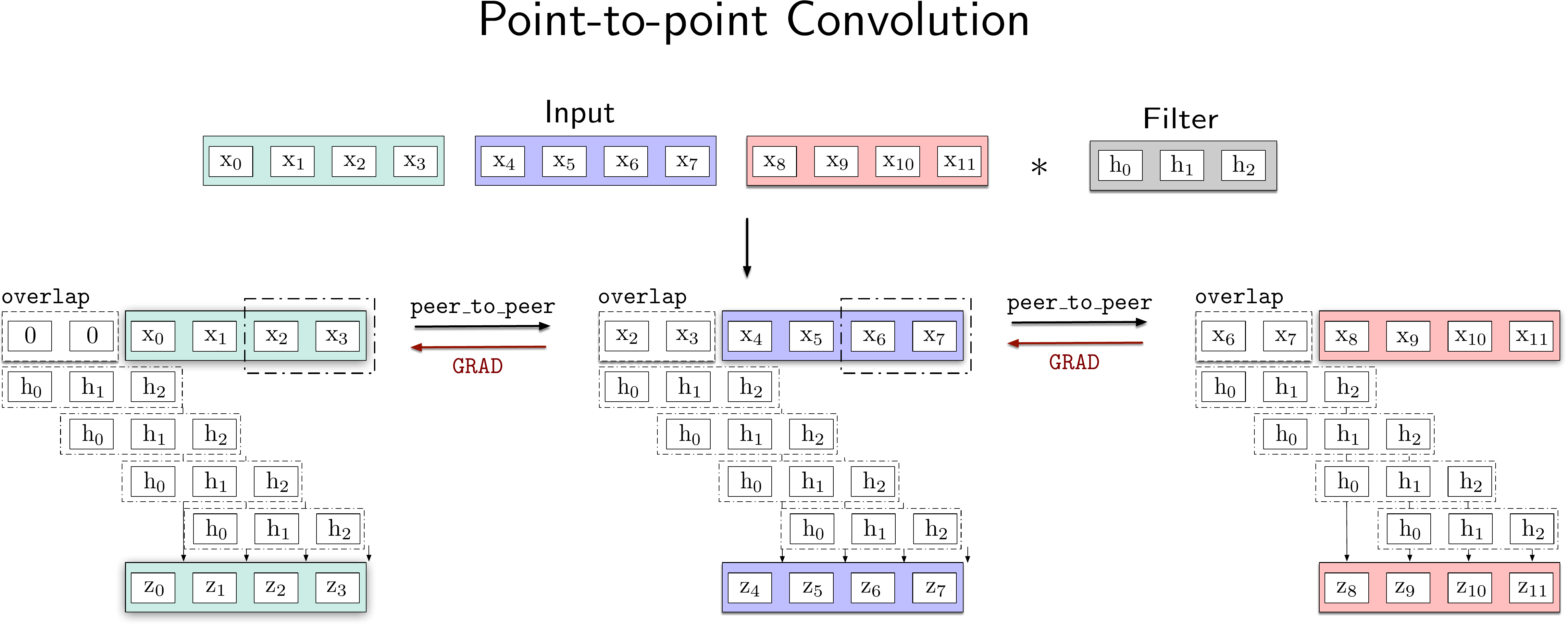}
    \caption{Diagram of computation and communication in \texttt{point-to-point convolutions}. This approach is best suited for FIR convolutions in \textsf{Hyena-SE} and \textsf{Hyena-MR}.}
    \label{fig:p2p_convs}
\end{figure}

\paragraph{{\tt Point-to-point} convolutions (Fig.~\ref{fig:p2p_convs})} Analogous to the self-attention case, \texttt{p2p} context parallel causal depthwise convolutions implements context parallelism while using communication with a single peer at a time. Let us consider the usual sharded input, with each rank holding $[\tH, \tLoverNcp]$, to be causally convolved with a filter $[\tH, \tK]$. We detail the depthwise case, but the grouped depthwise case can be obtained as a simplification. For FIR filters, one can exploit locality of the operation to simplify the implementation compared to {\tt p2p} attention. In particular, the causal convolution can be performed locally on most of the input elements without the need for communication. Only the first $\mathtt{\ell_h{-}1}$ elements on each shard rely on computation to be performed on a different rank, namely the last $\mathtt{\ell_h{-}1}$ elements of the previous shard. Note that each rank keeps or materializes copies of the convolutional filters, since each rank is responsible for computing convolutions across all $\tH$ channels. This is in contrast to {\tt a2a} schemes, which have ranks operate on chunks of channels independently. Figure \ref{fig:p2p_convs} provides a schematic of the algorithm.

\begin{note}{colback=blue!5}
    \textbf{[Extension] \texttt{Point-to-point} convolutions with overlapping communication:} Since only the first $\mathtt{\ell_h - 1}$ elements of each shard require inputs located on a different rank, we overlap local operations and \texttt{p2p} communication to further improve utilization. This process is illustrated in Supplementary Figure ~\ref{sup-fig:overlapping_comm}. Instead of waiting for the overlap segment to arrive before computing the convolution, we start the local convolution with a zero-padded input and simultaneously start the communication of the previous segment. Once communication is concluded, an additional convolution over the right-padded overlap of shape $[\tOne, \tH, \mathtt{2}\mathtt{(\ell_h{-}1)}]$ and the convolutional kernel is performed. The result of this convolution is subsequently added to the first $\mathtt{\ell_h - 1}$ elements of the previous output. Interestingly, this algorithm relies on similar decomposition techniques as those involved in our two-stage block convolution approach (Section \ref{sec:kernel_optimizations}). 
\end{note}

\begin{note}{colback=blue!5}
\textbf{[Extension] \texttt{Point-to-point} FFT Convolutions:}
While the previous CP algorithms can also be used for \textsf{Hyena-LI}, convolutions with long filters are generally implemented via Fast Fourier Transform (FFT) algorithms. FFT convolution relies on the fact that convolution is equivalent to multiplication in the Fourier domain:
\begin{equation}
    (x * h) = \mathsf{F}^{-1}(\mathsf{F}(x) \circ \mathsf{F}(h)), \label{eq:fft_conv}
\end{equation} where $\mathsf{F}, \mathsf{F}^{-1}$ are the Fourier and inverse Fourier transform, respectively. 

FFTs require access to the entire input sequence. At first glance, one could think it mandatory to host the whole sequence in a single rank. Interestingly, it is possible to compute both the Fourier and inverse Fourier transform --and thus the FFT convolution-- in a $\mathtt{p2p}$ fashion without ever hosting the whole sequence on a single device, by introducing communication during iterative steps of the FFTs.  Unfortunately, without further optimizations, we generally observed {\tt a2a} approaches to be faster for \textsf{Hyena-LI}. For completeness, we report the derivation in Appendix \ref{appx:p2p_fftconv}).

\end{note}

%% file: sections/discussion.tex
\section{Conclusion}

In this paper, we introduce systems and algorithms for convolutional multi-hybrids, a new class of architecture for sequence modeling at scale. We discuss architecture design, block layout, kernels for fast convolutions on GPUs based on overlap-add schemes, and context parallelism strategies. Multi-hybrids excel at efficient modeling of byte and character-level data, and we expect their utilization to unlock new applications for foundation models. Effectiveness of StripedHyena 2 is verified at scale (40 billion parameter, over 9 trillion tokens and 1 million context) with the Evo 2 line of models.

\section*{Acknowledgements}

We thank Armin W. Thomas and Keshigeyan Chandrasegaran for helpful discussions and feedback. 






%% file: sections/appendix/A_details.tex
\section*{Author Contributions}
\noindent

M.P. conceptualized the research; M.P. implemented the first version of the training infrastructure; J.K., E.N., D.W.R., G.Bri., B.Y., A.T., D.P.B., G.Bro., B.L.H., M.P., contributed to pretraining infrastructure; J.K., E.N., G.Bri., and M.P. designed pretraining experiments; M.P., E.N., designed the architectures; M.P. derived and implemented the first version of the two-stage block kernel; J.K. optimized the kernel and wrote the backward pass; S.M., M.P., extended the theory in section 3; D.W.R., B.Y., M.P., derived and implemented context parallelism for hyena and attention layers; D.W.R., derived and implemented the point-to-point scheme for spatial and FFT convolutions; M.P., G.Bri., A.V., wrote and optimized the inference stack; A.X.L., contributed to inference stack; C.R., P.D.H., B.L.H., S.E., M.P., supervised the project.

\section{Appendix: Additional Results}

\subsection{Additional Algorithms for Direct Convolution on Tensor Cores}\label{sec:two_stage_algo_tensor_cores}

Modern GPU accelerators include \emph{tensor cores} capable of high-throughput dense matrix multiplications (\texttt{GEMM}). In the context of the two-stage block algorithm described in Section~\ref{sec:two_stage_algo}, an effective strategy is to recast small matrix-vector products into larger \texttt{GEMM} kernels that better exploit these tensor units. Below, we focus on a \emph{single group of $d_g$ channels}, which all share one depthwise FIR filter of length $\ell_h$. We assume $\ell_h \le 2\,\ell_b$, so only two $(\ell_b\times \ell_b)$ Toeplitz blocks are needed for the filter:
\[
    H_0, \quad H_1 \;\in\; \RR^{\ell_b \times \ell_b}.
\]
These blocks respectively handle the “current-chunk” taps and the “spillover” taps from the preceding chunk. We provide here a different algorithm to transform direct convolutions into {\tt GEMMs}, even without grouping. The idea is to parallelize across the chunks, rather than channels.

\paragraph{Block decomposition and dimensions.} Let $\hat{X}_n \;\in\; \RR^{\ell_b \times d_g}$, $\hat{Y}_n \;\in\; \RR^{\ell_b \times d_g}$
denote the $n$-th input and output blocks (or \emph{chunks}) for the group. Here, $\ell_b$ is the block size along the time dimension, and $d_g$ is the number of channels in the group. According to the two-stage block convolution framework (cf.\ Section~\ref{sec:two_stage_algo}), each output block is given by
\begin{equation}
\label{eq:two_stage_tensor_core_single_group}
    \hat{Y}_n 
    \;=\;
    H_0 \,\hat{X}_n
    \;+\;
    H_1 \,\hat{X}_{n-1},
    \qquad
    n \;=\; 0,1,\dots,\bigl\lceil \tfrac{\ell}{\ell_b} \bigr\rceil - 1,
\end{equation}
with the convention that $\hat{X}_{-1} = 0$ (i.e., the “previous chunk” is zero for $n=0$). In particular, $H_0$ and $H_1$ are constant for all chunks $n$ once the filter has been fixed. Differently from the grouped algorithm proposed in the main text, this approach parallelizes across chunks, requiring the a reshape on the input.

\paragraph{Mapping to tensor cores.}
Equation \eqref{eq:two_stage_tensor_core_single_group} naturally translates into two matrix-matrix multiplications plus an elementwise addition:
\[
    \hat{Y}_n
    \;=\;
    \underbrace{H_0\,\hat{X}_n}_{(\ell_b\times \ell_b)\,\times\,(\ell_b\times d_g)}
    \;+\;
    \underbrace{H_1\,\hat{X}_{n-1}}_{(\ell_b\times \ell_b)\,\times\,(\ell_b\times d_g)}.
\]
Because $H_0$ and $H_1$ remain unchanged for all $n$, the following procedure can be used to implement \eqref{eq:two_stage_tensor_core_single_group} efficiently on GPUs:
\begin{enumerate}
    \item \textbf{Filter preload}  
    Load $H_0$ and $H_1$ from global memory into low-latency on-chip memory (e.g., shared memory). This is a one-time overhead amortized across all chunks.
    \item \textbf{Chunk read}  
    Read the current chunk $\hat{X}_n$ (and the previous chunk $\hat{X}_{n-1}$ if $n>0$) from global memory into local registers or shared memory. 
    \item \textbf{Tensor-core \texttt{GEMM}}
    Perform the matrix multiplications $H_0 \,\hat{X}_n$ and $H_1 \,\hat{X}_{n-1}$ on the tensor cores, accumulating the two partial results into $\hat{Y}_n$.
    \item \textbf{Output writeback}  
    Write the output block $\hat{Y}_n$ to global memory.
\end{enumerate}
\paragraph{Cost model}
The floating-point cost per chunk follows directly from \eqref{eq:two_stage_tensor_core_single_group}. Each chunk output $\hat{Y}_n$ requires two matrix multiplications of dimension $(\ell_b \times \ell_b)$ by $(\ell_b \times d)$ for a total of $2\,\ell_b^2\,d$ floating-point operations. Summing over all $\lceil \ell/\ell_b \rceil$ chunks, the total floating-point operations per sequence (for a single group) is $2\,\ell_b^2\,d\,\lceil \ell/\ell_b \rceil$.



\subsection{Context Parallel Methods}

\subsubsection{All-to-All Attention}\label{sup-sec:ulysses}

In the case of self-attention, where channels ${\tt D}$ are split into groups (heads), and operations happen independently over each head, the sharding is done such that an integer number of heads is held on each device. Once the operation is complete, a second $\texttt{a2a}$  exchange is used to return to the input shape of $[\mathtt{1,H, L//N_{CP}}]$. {\tt a2a} context parallelism has been used in attention parallelization strategies, such as DeepSpeed Ulysses \citep{jacobs2023deepspeed}.

\begin{figure}[H]
    \centering
    \includegraphics[width=0.99\textwidth]{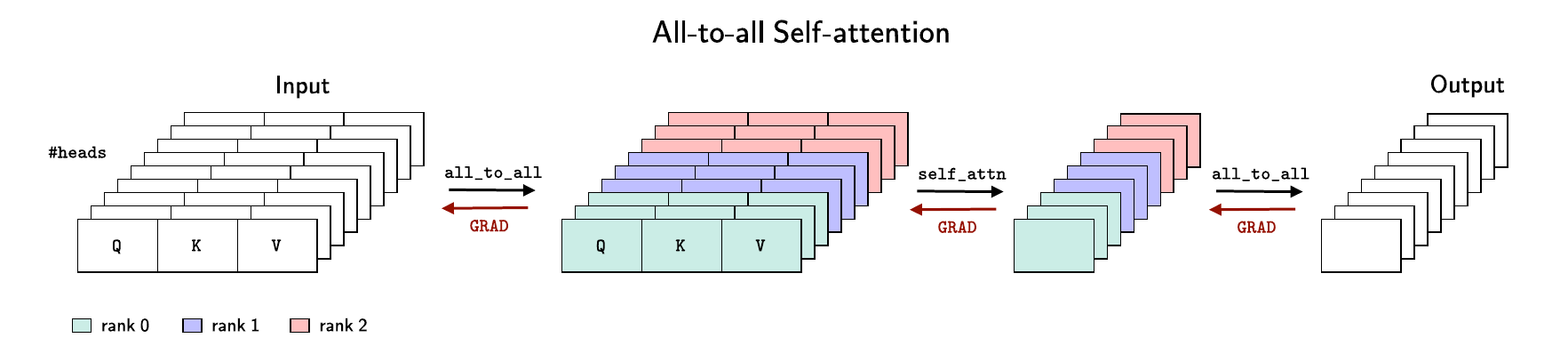}
    \caption{Diagram of computation and communication in \texttt{all-to-all self-attention}. Attention heads are split across devices such that an integer number of heads is held on each device. Then, self-attention is performed locally on each device and subsequently, the output is exchanged across all ranks to reconstruct the original shape of the input.}
    \label{fig:a2a_self_attn}
\end{figure}

\subsubsection{Point-to-Point Attention}\label{sup-sec:p2p-att}

Ring attention methods propose a $\texttt{p2p}$ solution for self-attention. For simplicity, let us assume that the query, key and value projections of the input are all of same size as the input shape $[\mathtt{\tH, \tLoverNcp}]$. Ring-Attention arranges CP ranks in a ring, each holding a portion of the query. Consequently, Ring Attention (RA) \citep{liu2023ring} passes portions of the key and value tensors, each of shape $[\mathtt{\tH, \tLoverNcp}]$ following the ring arrangement. At each stage, self-attention is performed between the query stationed in each device and the transmitted key-value portions. Global statistics are updated online on each device based on each upcoming key value portions to compute the softmax. After $\tNcp$ steps, each consisting of $\tNcp$ parallel $\mathtt{p2p}$ calls, the shards of the queries held on each device will have seen all key-value portions, and each device will hold a shard of the final attention result. Combining this result with the online softmax operation, full attention operation can be concluded without ever holding the whole sequence on a single device.

\begin{figure}[H]
    \centering
    \includegraphics[width=0.99\textwidth]{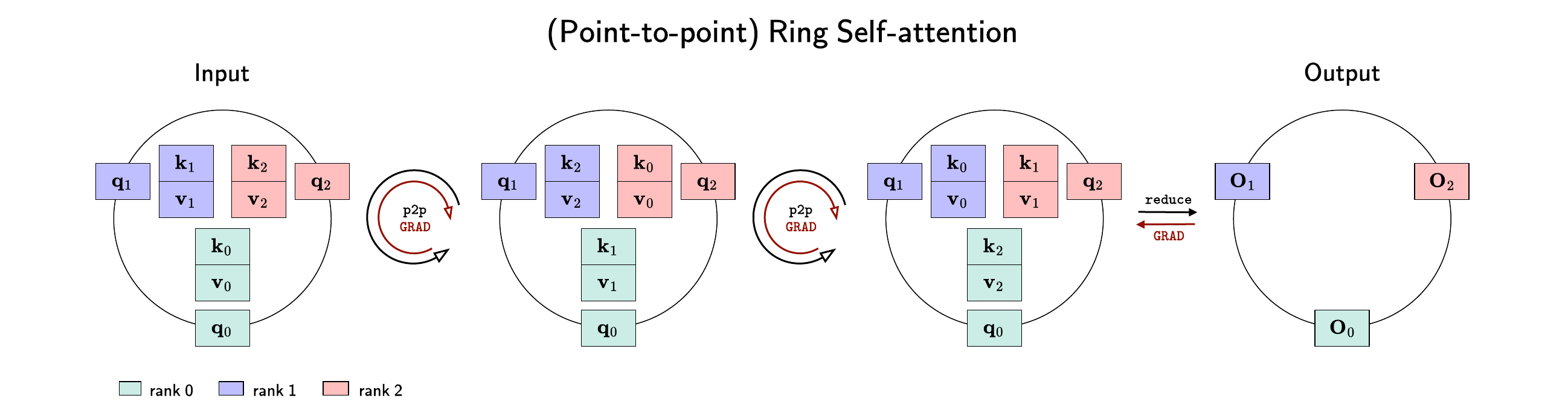}
    \caption{Diagram of computation and communication in \texttt{(point-to-point) ring self-attention}. The key and value chunks are communicated in a ring arrangement until all devices have seen all of the chunks. Once all chunks have been seen, partial results can be reduced to complete the attention operation.}
    \label{fig:p2p_ring_self_attn}
\end{figure}

\subsubsection{Considerations for Causal Models.}\label{sup-sec:causal} When causal transformers are used, e.g., for autoregressive tasks, computations follow a triangular structure. \citet{brandon2023striped} noticed that this triangular structure causes load imbalances in Ring-Attention in causal autoregressive settings. To overcome this problem, they propose to shard the input across CP devices using $2{\times}$ as many shards as CP ranks and accommodate $2$ shards on each CP rank using their introduced \textit{striped ordering}. For example, for $\{\vec{x}_{j}\}_{j=1}^{2 \tNcp}$ shards with $\tNcp{=}\mathtt{4}$, shards are organized as: $[\vec{x}_{0}, \vec{x}_{4}]$, $[\vec{x}_{1}, \vec{x}_{5}]$, $[\vec{x}_{2}, \vec{x}_{6}]$, $[\vec{x}_{3}, \vec{x}_{7}]$ on each rank respectively. \citet{dubey2024llama} proposed an improved load balancing strategy for the training of Llama-3 by distributed shards following a \textit{zig-zag splitting}. Specifically, for the same case with $\{\vec{x}_{j}\}_{j=1}^{2 \tNcp}$ shards and $\tNcp{=}\mathtt{4}$, shards are organized as: $[\vec{x}_{0}, \vec{x}_{7}]$, $[\vec{x}_{1}, \vec{x}_{6}]$, $[\vec{x}_{2}, \vec{x}_{5}]$, $[\vec{x}_{3}, \vec{x}_{4}]$ on each rank respectively.

For the training of StripedHyena 2, we adopt the zig-zag splitting of \citet{dubey2024llama}. We note that the choice of sharding strategy has no major implications for any of the \texttt{a2a} and \texttt{p2p} CP strategies introduced before. For \texttt{a2a}, the zigzag arrangement must now be considered when reconstructing the sequence. For \texttt{p2p}, two buffers must be kept on each rank, each representing the attention operation on each split.



\subsubsection{Point-to-point FFT Convolutions}\label{appx:p2p_fftconv}

As mentioned in the main text, the idea behind \texttt{p2p} Context Parallelism is to perform an operation over an input of shape [$\tOne, \tH, \tL$] sharded along the sequence dimension onto $\tNcp$ devices, each holding a shard of shape [$\tOne, \tH, \tLoverNcp$], without ever holding the whole sequence on a single device. 


This section heavily relies on the FFT derivation and the multiple Radix-$N$ algorithms presented in \citet{takahashi2019fast}. We refer interested readers to that textbook for additional details.

\paragraph{Primer on the Fast Fourier Transform.} 

The name Fast Fourier Transform (FFT) comes from an algorithm that, as its name indicates, allows computing the Discrete Fourier Transform fast. Computing the Discrete Fourier Transform naively has quadratic complexity. However, the FFT is able to achieve the same result with $\mathcal{O}(\tL \log \tL)$ complexity, by using a divide and conquer formulation. 

To understand the FFT, we start from the Discrete Fourier Transform (DFT) defined as:
\begin{equation}
y(k) = \mathrm{DFT}_l(x) = \sum_{j=0}^{l-1} x(j)\omega_l^{jk}, \quad 0 \leq k \leq l-1,
\label{eq:dft}
\end{equation}
where \(\omega_l {=} e^{-2\pi i / l}\) and \(i {=} \sqrt{-1}\), applied to an input $x$ of length $l{=}4$. The DFT of $x$ can be calculated as:
\begin{equation}
\begin{bmatrix}
y(0) \\ 
y(1) \\ 
y(2) \\ 
y(3)
\end{bmatrix}
=
\begin{bmatrix}
\omega^0 & \omega^0 & \omega^0 & \omega^0 \\ 
\omega^0 & \omega^1 & \omega^2 & \omega^3 \\ 
\omega^0 & \omega^2 & \omega^4 & \omega^6 \\ 
\omega^0 & \omega^3 & \omega^6 & \omega^9
\end{bmatrix}
\begin{bmatrix}
x(0) \\ 
x(1) \\ 
x(2) \\ 
x(3)
\end{bmatrix}.
\label{eq:second_step}
\end{equation}
Importantly, the terms $\omega_l^{jk}$ are not independent. In fact, there is a relation $\omega_l^{jk} {=} \omega_l^{jk \bmod l}$, which we can use to rewrite Eq.~\ref{eq:second_step} as follows:
\begin{equation}
\begin{bmatrix}
y(0) \\ 
y(1) \\ 
y(2) \\ 
y(3)
\end{bmatrix}
=
\begin{bmatrix}
1 & 1 & 1 & 1 \\ 
1 & \omega^1 & \omega^2 & \omega^3 \\ 
1 & \omega^2 & \omega^0 & \omega^2 \\ 
1 & \omega^3 & \omega^2 & \omega^1
\end{bmatrix}
\begin{bmatrix}
x(0) \\ 
x(1) \\ 
x(2) \\ 
x(3)
\end{bmatrix}.
\end{equation}
At this point, we can observe that there are several values that repeat themselves. Using some algebra and reorganizing the position of the output positions in the vector, we arrive at the following decomposition:
\begin{equation}
\begin{bmatrix}
y(0) \\ 
y(2) \\ 
y(1) \\ 
y(3)
\end{bmatrix}
=
\begin{bmatrix}
1 & \omega^0 & 0 & 0 \\ 
1 & \omega^2 & 0 & 0 \\ 
0 & 0 & 1 & \omega^1 \\ 
0 & 0 & 1 & \omega^3
\end{bmatrix}
\begin{bmatrix}
1 & 0 & \omega^0 & 0 \\ 
0 & 1 & 0 & \omega^0 \\ 
1 & 0 & \omega^2 & 0 \\ 
0 & 1 & 0 & \omega^2
\end{bmatrix}
\begin{bmatrix}
x(0) \\ 
x(1) \\ 
x(2) \\ 
x(3)
\end{bmatrix}.
\label{eq:final_matrix_form_example}
\end{equation}
Looking at this decomposition, we observe two important things: First, the first of the matrices is a blockwise matrix, corresponding to two DFTs of half the size of the initial sequence length. In the general case, it holds that, an $l$-point DFT can be decomposed onto two $\frac{l}{2}$-point DFTs followed by some arithmetic operations.\footnote{While this observation is already enough for us to describe \texttt{p2p} FFT convolutions, it is worth noticing that the FFT repeats this process until the input can no longer be split, i.e., $l{=}2$. This recursive split procedure is that makes the FFT fast.} Specifically, for an input $x$ of length $l$, divided onto two chunks $x(j)$, $x(j + l / 2)$, its DFT is given by:
\begin{equation}
\begin{aligned}
y(k) &= \mathrm{DFT}_{l/2}\big(x(j) + x(j + l/2)\big) \\
y(k + 1) &= \mathrm{DFT}_{l/2}\left(\omega^j_l (x(j) - x(j + l/2))\right).
\end{aligned}
\label{eq:fft_decomposition}
\end{equation}
Secondly, it is important to note that while the values of $x$ are organized sequentially in Eq.~\ref{eq:final_matrix_form_example}, the values of its DFT $y$ have been permuted following a bit reversal order. In the general case, there exist two types of FFT depending on whether the input is assumed to be organized sequentially --in which case the output is bit reversed--, or whether the input is assumed to be bit reversed --in which case the output is organized sequentially. These are known as \textit{Decimation-in-Frequency} (DiF) and \textit{Decimation-in-Time} (DiT) FFT algorithms, respectively. Equation \ref{eq:fft_decomposition} depicts the DiF FFT algorithm.

\paragraph{Butterfly diagrams.} A powerful, intuitive way to visualize the flow of data in DiF and DiT FFTs are butterfly diagrams. Butterfly diagrams illustrate the data exchange in DiF and DiT FFTs, which are based on two operations:

\par\vspace{-1em} 

\noindent
\begin{minipage}[t]{0.45\textwidth}
\begin{equation}\label{eq:butterfly:DiF}
\begin{aligned}
&\textbf{DiF Butterfly:}\\
&X = X + Y, \\
&Y = (X - Y) \ \omega^j \\
\end{aligned}
\end{equation}
\end{minipage}
\begin{minipage}[t]{0.45\textwidth}
\begin{equation}\label{eq:butterfly:DiT}
\begin{aligned}
&\textbf{DiT Butterfly:}\\
&X = X + \omega^j\ Y, \\
&Y = X - \omega^j\  Y  \\
\end{aligned}
\end{equation}
\end{minipage}

\begin{figure}[t]
  \centering
  \begin{subfigure}[b]{0.45\textwidth}
    \includegraphics[width=\textwidth]{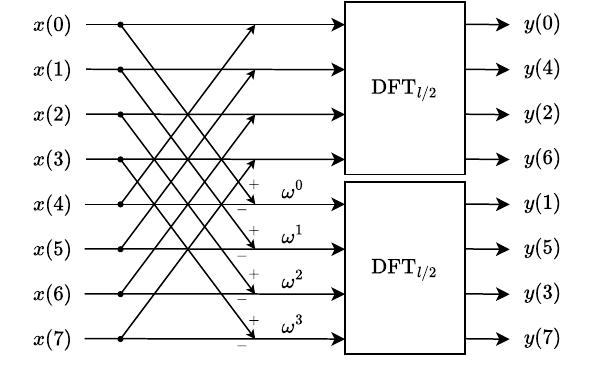}
    \caption{DiF-FFT butterfly diagram}
    \label{fig:dit}
  \end{subfigure}
  \hfill
  \begin{subfigure}[b]{0.45\textwidth}
    \includegraphics[width=\textwidth]{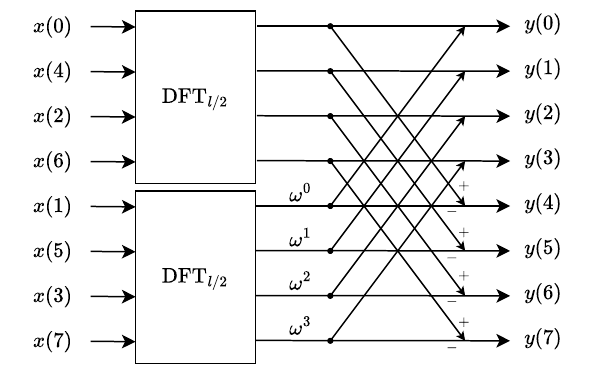}
    \caption{DiT-FFT butterfly diagram}
    \label{fig:dif}
  \end{subfigure}
  \caption{Butterfly diagrams for DiT and DiF FFTs.}
  \label{fig:butterfly}
\end{figure}

The butterfly diagrams of DiF and DiT Fast Fourier Transforms are shown in Fig.~\ref{fig:butterfly}.

\paragraph{The inverse DFT.} The inverse DFT (iDFT) is defined as:
\begin{equation}
x(j) = \mathrm{iDFT}_l(y) = \frac{1}{l}\sum_{k=0}^{l-1} y(k)\omega_l^{-jk}, \quad 0 \leq j \leq l-1.
\end{equation}
When compared to the DFT (Eq.~\ref{eq:dft}), we observe that the only differences are (i) the normalization factor $\frac{1}{l}$, and (ii) the minus sign in the $\omega^{-jk}$. Luckily, this means that the exact same operation as well as butterfly diagrams can be used, with two minor modifications: all $\omega^j$ terms are exchanged by an $\omega^{-j}$ term, and the normalization by $\frac{1}{l}$ must be considered.  

\subsubsection{Point-to-point FFT Convolutions with CP{=}2}

Looking at the FFT formulation, we can observe that the FFT is computed by performing independent FFTs on two independent splits of the input, followed by some point-wise operations over these splits. Interestingly, this setting exactly resembles the case of a distributed \texttt{p2p} setting, where each split is held in a different device. Nonetheless, there is an important impediment due to the organization of the data \textit{after} the FFT is performed. Assuming the conventional sequential splitting of the input across CP ranks, once a distributed FFT is performed, the output of the FFT would be bit-reversed across all ranks. Consequently, in order to restore the sharding distribution of the input, an additional \texttt{a2a} call would be required.

Luckily, this permutation of the output is not a problem when performing FFT Convolutions. Since the FFT convolution is composed of an FFT followed by an inverse FFT, it is sufficient to use a forward (DiF) FFT --that generates a bit-reversed output-- followed by a DiF inverse FFT algorithm to generate an output that follows the same organization as the input. As a result, after the FFT convolution is finished, both the input and the output will be sharded in the same manner.

Listing~\ref{listing:minimal_fftconv} shows a minimalist reference implementation of the \texttt{p2p} FFT convolution with simulated sharding for $\tNcp{=}2$.


\begin{listing}
\begin{minted}{python}
def bit_reversal(xarray: torch.Tensor, size: int, log2length: int):
    """Applies a bit-reversal permutation to the input array."""
    reversed_indices = vectorized_bit_reversal_indices(size, log2length, xarray.device)
    return xarray[..., reversed_indices]

def dif_radix2_fft(x: torch.Tensor):
    """Applies the radix-2 Decimation-in-Frequency (DIF) FFT to the input.

    Parameters:
    - x (torch.Tensor): The input tensor of shape [B, H, L]

    Returns:
    - tuple(torch.Tensor, torch.Tensor): The sharded bit-reversed fft of the input.
    """
    # Split the input (to simulate a CP group of size 2).
    _x_0, _x_1 = rearrange(x,"... (n1 n2) -> ... n1 n2", n1 = 2, n2 = N // 2).unbind(dim=-2)

    # Twiddle factors
    k = torch.arange(N // 2, device=x.device)
    W = torch.exp(-1j * 2.0 * torch.pi * k / N)

    # Apply butterfly operations
    x_0 = _x_0 + _x_1
    x_1 = (_x_0 - _x_1) * W

    # Compute FFT on both halves.
    fft_x_0 = torch.fft.fft(x_0, dim=-1), N // 2, int(math.log2(N // 2)))
    fft_x_1 = torch.fft.fft(x_1, dim=-1), N // 2, int(math.log2(N // 2)))
    return fft_x_0, fft_x_1

def dif_radix2_ifft(fft_x_0, fft_x_1: torch.Tensor):
    """Applies the radix-2 iFFT assuming that the input is a bit-reversed FFT, and returns 
       a non-reversed input tensor, i.e., x = dif_radix2_ifft(dif_radix2_fft(x))."""

    # Compute iFFT on both halves (it internally performs normalization by 1 / n // 2). 
    _x_0 = torch.fft.ifft(fft_x_0, dim=-1)
    _x_1 = torch.fft.ifft(fft_x_1, dim=-1)

    # Twiddle factors
    k = torch.arange(N // 2, device=_x_1.device)
    W = torch.exp(1.j * 2.0 * torch.pi * k / N)

    # Apply butterfly operations
    x_0 = _x_0 + W * _x_1
    x_1 = _x_0 - W * _x_1

    # Normalize by seq. length of this stage (2) & concat results for verification.
    return 0.5 * torch.cat([x_0, x_1], dim=-1)

assert torch.allclose(x, dif_radix2_ifft(dif_radix2_fft(x)).real)
\end{minted}
\caption{Minimal implementation of the FFT and iFFT with a simulated CP group of size \texttt{2}.}
\label{listing:minimal_fftconv}
\end{listing}

\subsection{Extending \texttt{p2p} FFT Convolutions to larger CP sizes} 
The previous section illustrates how a \texttt{p2p} FFT convolution can be computed for a CP group with $\tNcp{=}2$ devices. In this section, we show how to extend this procedure to CP groups with $\tNcp{>}2$ devices.

\paragraph{Radix-$\tN$ FFT.} Before we continue, we must introduce the concept of a Radix-$\tN$ FFT algorithm. Simply put, a Radix-$\tN$ FFT algorithm computes a $l$-point FFT by decomposing it onto $\tN$ independent $l \mathtt{//} \tN$-point FFTs followed by pointwise operations. In other words, a Radix-$\tN$ FFT algorithm generalizes the splitting process illustrated in the previous section for a value of $\tN{=}2$ to values $\tN{>}2$. Furthermore, just as for the Radix-\texttt{2} FFT algorithm, there exist DiT and DiF implementations for several values of \texttt{N}. \citet{takahashi2019fast} provides an exceptional description of Radix-$\tN$ FFT algorithms for $\tN\in[\mathtt{3},\mathtt{4},\mathtt{5},\mathtt{8}]$, and many algorithms exist for many other values of $\tN$, e.g., $\tN{=}\mathtt{16}$ \citep{bouguezel2004improved, huang2012high}. An important difference from the $\tN{=}\mathtt{2}$ case, is that the Radix-$\tN$ algorithms introduces $\tN$ different sets of twiddle factors. For $\tN{=}\mathtt{2}$, we had two sets of values $W_{0j}=[\omega_{0\times0}, ... , \omega_{(\frac{l}{2}-1)\times0}] = [1, ..., 1]$, and $W_{1j}=[\omega_{0\times1}, ... , \omega_{(\frac{l}{2} -1)\times1}] = [1, \omega^1, \omega^2, ..., \omega^{\frac{l}{2}}]$ applied to the first and second splits, respectively (Fig.~\ref{fig:butterfly}). For a general value of $\tN$, we utilize $\tN$ sets of twiddle factors, $\{W_{nj}\}_{n=0}^{\tN-1}$, defined as $W_{nj}=[\omega_{0\times n}, ..., \omega_{(\frac{l}{\tN}-1)\times n}]$.

\begin{figure}[t]
  \centering
  \begin{subfigure}[b]{0.49\textwidth}
    \includegraphics[width=1\textwidth]{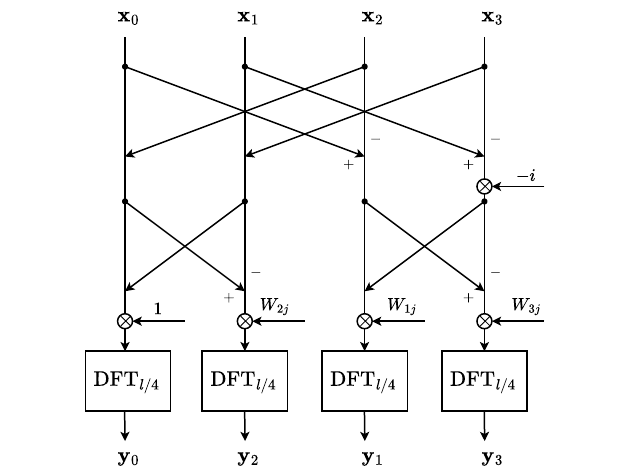}
    \caption{\texttt{p2p} FFT for $\tNcp{=}4$ devices.}
    \label{fig:fftconv_cp4_fft}
  \end{subfigure}
  \begin{subfigure}[b]{0.49\textwidth}
    \includegraphics[width=1\textwidth]{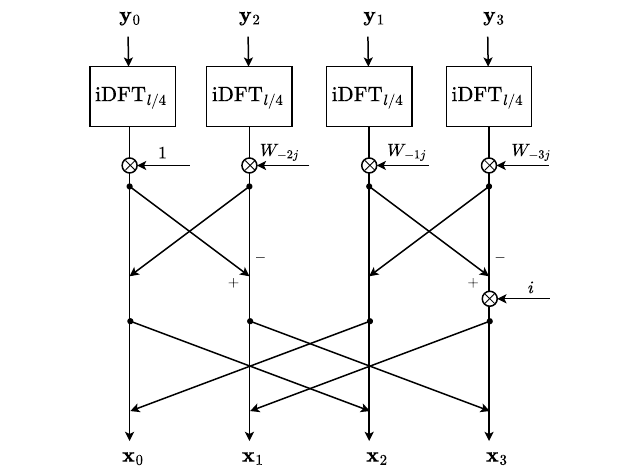}
    \caption{\texttt{p2p} inverse FFT for $\tNcp{=}4$ devices.}
    \label{fig:fftconv_cp4_ifft}
  \end{subfigure}
  \caption{Butterfly diagrams for the distributed \texttt{p2p} FFT and  inverse FFT for $\tNcp{=}4$. $\boldsymbol{\mathrm{x}}_{0}, ..., \boldsymbol{\mathrm{x}}_{3}$ represent the shards of the input $\boldsymbol{\mathrm{x}}$ held on each CP rank. Devices are represented by the vertical lines. Note that after the FFT, the outputs $\{\boldsymbol{\mathrm{y}}_{j}\}_{j=0}^3$ are bit-reversed over CP ranks. However, after combining it with the inverse FFT, the same sharding distribution as the input is obtained. FFT convolutions are computed by performing the FFT of both the input and the (sharded) convolutional kernel, multiplying them in the Fourier domain, and returning the result back to the spatial domain.}
  \label{fig:fftconv_cp4}
\end{figure}

\begin{figure}[H]
  \centering
  \begin{subfigure}[b]{0.8\textwidth}
    \includegraphics[width=1\textwidth]{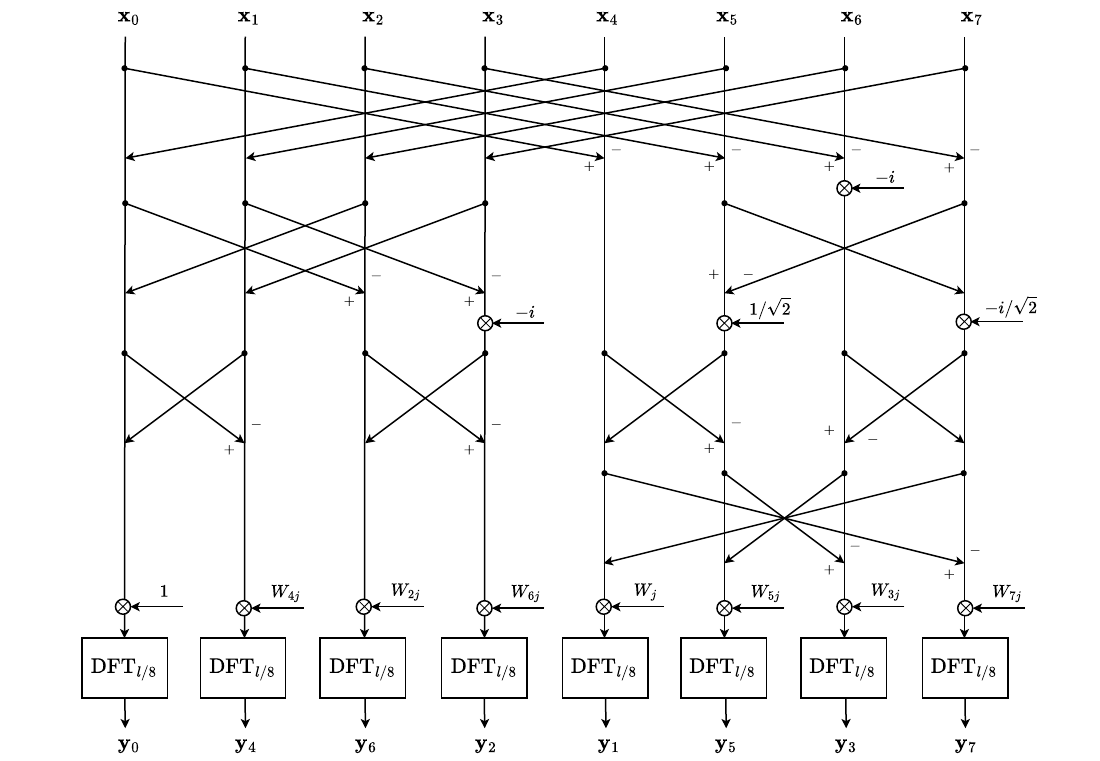}
    \caption{\texttt{p2p} FFT for $\tNcp{=}\mathtt{8}$ devices.
    \vspace{4mm}}
    \label{fig:fftconv_cp8_fft}
  \end{subfigure}
  \begin{subfigure}[b]{0.8\textwidth}
    \includegraphics[width=1\textwidth]{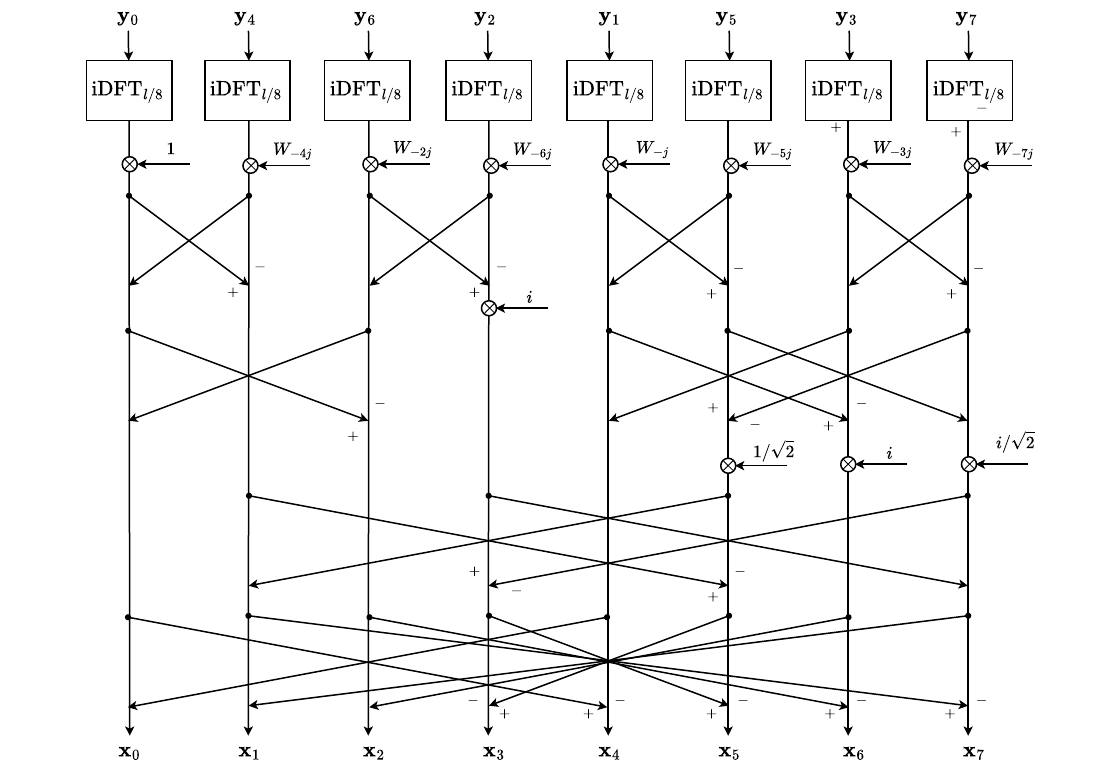}
    \caption{\texttt{p2p} inverse FFT for $\tNcp{=}\mathtt{8}$ devices.}
    \label{fig:fftconv_cp8_ifft}
  \end{subfigure}
  \caption{Butterfly diagrams for the distributed \texttt{p2p} FFT and  inverse FFT for $\tNcp{=}\mathtt{8}$. $\boldsymbol{\mathrm{x}}_{0}, ..., \boldsymbol{\mathrm{x}}_{7}$ represent the shards of the input $\boldsymbol{\mathrm{x}}$ held on each CP rank. Devices are represented by the vertical lines. Note that after the FFT, the outputs $\{\boldsymbol{\mathrm{y}}_{j}\}_{j=0}^7$ are bit-reversed over CP ranks. However, after combining it with the inverse FFT, the same sharding distribution as the input is obtained. FFT convolutions are computed by performing the FFT of both the input and the (sharded) convolutional kernel, multiplying them in the Fourier domain, and returning the result back to the spatial domain.}
  \label{fig:fftconv_cp8}
\end{figure}

\paragraph{Extension to $\tNcp{>}2$ devices.} Following the same formulation used for $\tN{=}2$, we can compose a DiF Radix-$\tN$ FFT and an inverse DiF Radix-$\tN$ FFT to perform convolutions in a distributed setting. Specifically, given an input of shape $[1, \tH, \tL]$ sharded on a CP group with $\tNcp$ devices, the \texttt{p2p} FFT convolution is implemented by using Radix-$\tNcp$ (DiF) FFT and inverse (DiF) FFT algorithms to compute the FFT convolution without holding the whole sequence in a single device. Schematic butterfly diagrams for the implementation of \texttt{p2p} FFT convolutions for $\tNcp{=}\mathtt{4}$, and $\tNcp{=}\mathtt{8}$ devices are provided in Figs.~\ref{fig:fftconv_cp4},~\ref{fig:fftconv_cp8}.

\subsection{Two-Pass Algorithm}\label{sup-sec:kernel}

\textbf{Backward kernel}: To compute filter gradients in the backward pass, one requires global accumulation. Instead of limiting the computation to a single kernel, we implement gradient calculation using back-to-back kernels. The first performs a partial accumulation of the filter gradient by block, maintaining the same overall structure as the forward kernel, while the second kernel calculates the final result as a reduction of these partial gradients. Importantly, we take care to write out the partially accumulated gradients in coalesced format to enable a simple vectorized reduction as a second step.  

\paragraph{Fast materialization of Toeplitz factors}
In the listing below, we report code to efficiently materialize the Toeplitz factors using Triton. 

\begin{listing}
\begin{minted}{python}

import triton
import triton.language as tl

@triton.jit
def toeplitz_idx(
    FILTER_LEN: tl.constexpr,
    CHUNK_SIZE: tl.constexpr,
    TOEPLITZ_TYPE: tl.constexpr = "toeplitz",
):

    if TOEPLITZ_TYPE == "zeroth_factor":
        r_idx = tl.arange((FILTER_LEN - 1), CHUNK_SIZE + (FILTER_LEN - 1))[None, :]
    elif TOEPLITZ_TYPE == "first_factor":
        r_idx = (
            tl.arange((FILTER_LEN - 1), CHUNK_SIZE + (FILTER_LEN - 1))[None, :]
            - CHUNK_SIZE
        )
    else:
        tl.static_assert(False, "Invalid ToeplitzType")
    c_idx = tl.arange(0, CHUNK_SIZE)[:, None]
    idx = r_idx - c_idx
    return idx

@triton.jit
def load_toeplitz(
    h_ptr,
    FILTER_LEN: tl.constexpr,
    CHUNK_SIZE: tl.constexpr,
):
    t_idx = toeplitz_idx(FILTER_LEN, CHUNK_SIZE, "toeplitz")
    mask = (0 <= t_idx) & (t_idx < FILTER_LEN)

    T = tl.load(
        h_ptr + group_num * FILTER_LEN + t_idx,
        mask=mask,
        other=0.0,
        eviction_policy="evict_last",
    )

    return T
\end{minted}
\caption{Masked loading of Toeplitz matrix factors $T_h^{(0)}$ and $T_h^{(1)}$ in Triton.}
\end{listing}

%% file: sections/appendix/B_figures.tex
\section{Appendix: Supplementary Figures}

\begin{figure}[H]
    \centering
    \includegraphics[width=0.5\textwidth]{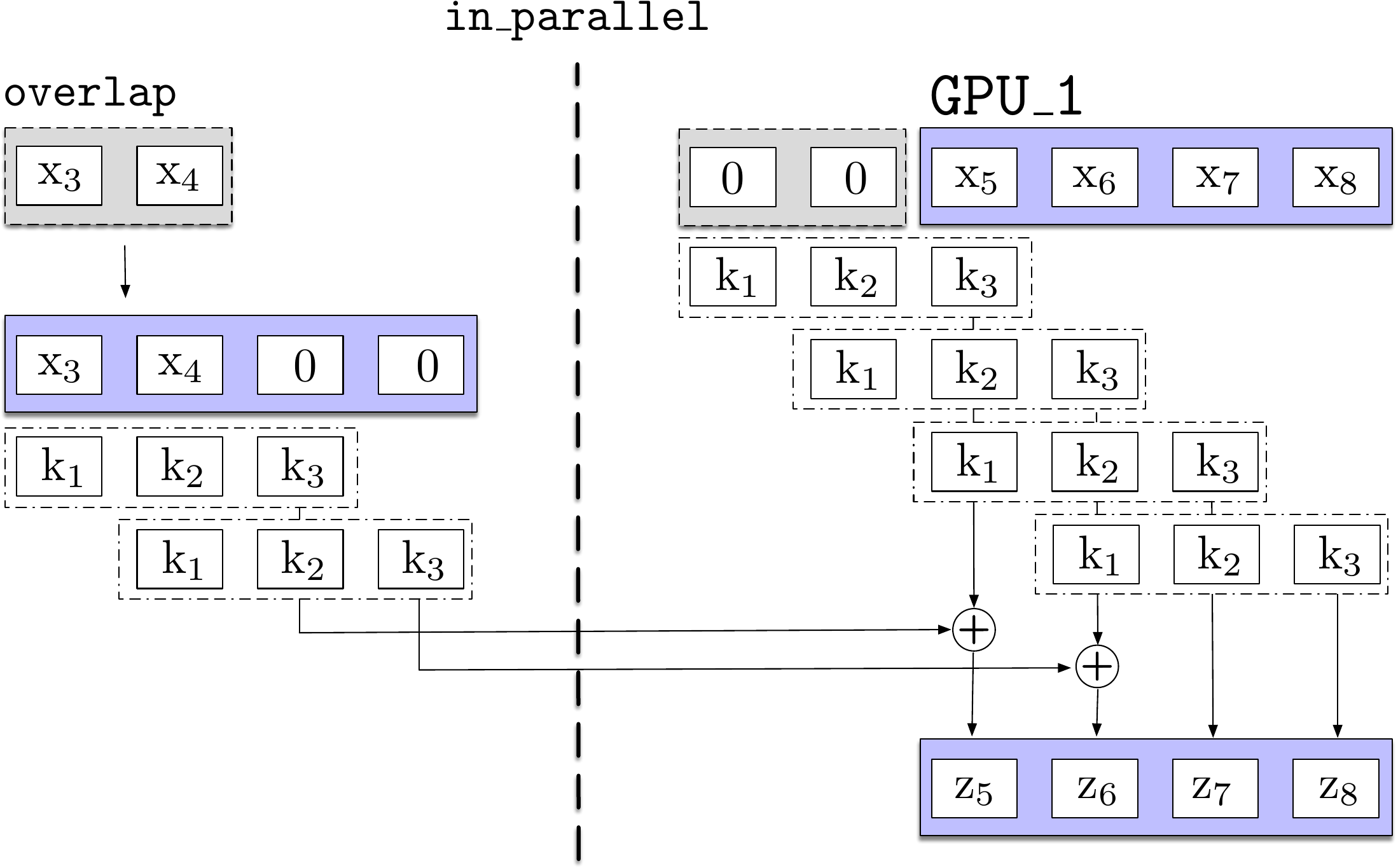}
    \caption{Diagram of computation in our overlapped {\tt point-to-point convolution} scheme.}
    \label{sup-fig:overlapping_comm}
\end{figure}

\begin{figure}[H]
    \includegraphics[width=0.99\textwidth]{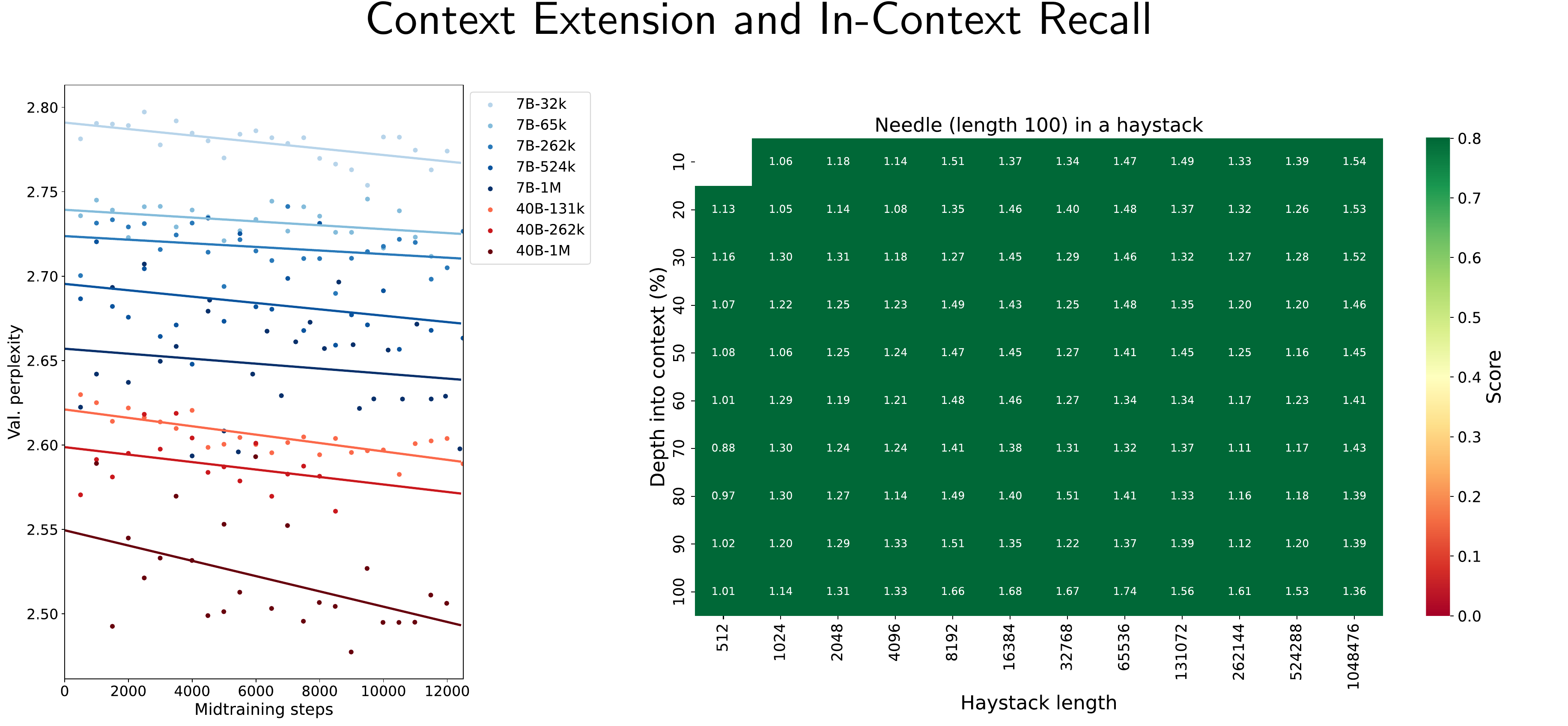}
    \vspace{-0.1cm}
    \caption{\textbf{[Left]:} Validation perplexity on {\tt OpenGenome2} after midtraining extension with different techniques, at model scales of 7B and 40B. The extensions are performed on the base Evo 2 7B and 40B models. We also provide a linear fit at each scale. \textbf{[Right]:} Recall performance of 40B 1M measured via the needle-in-the-haystack task described in \citep{brixievo2}}
    \label{fig:niah_extension}
\end{figure}

\begin{figure}[H]
    \includegraphics[width=0.45\textwidth]{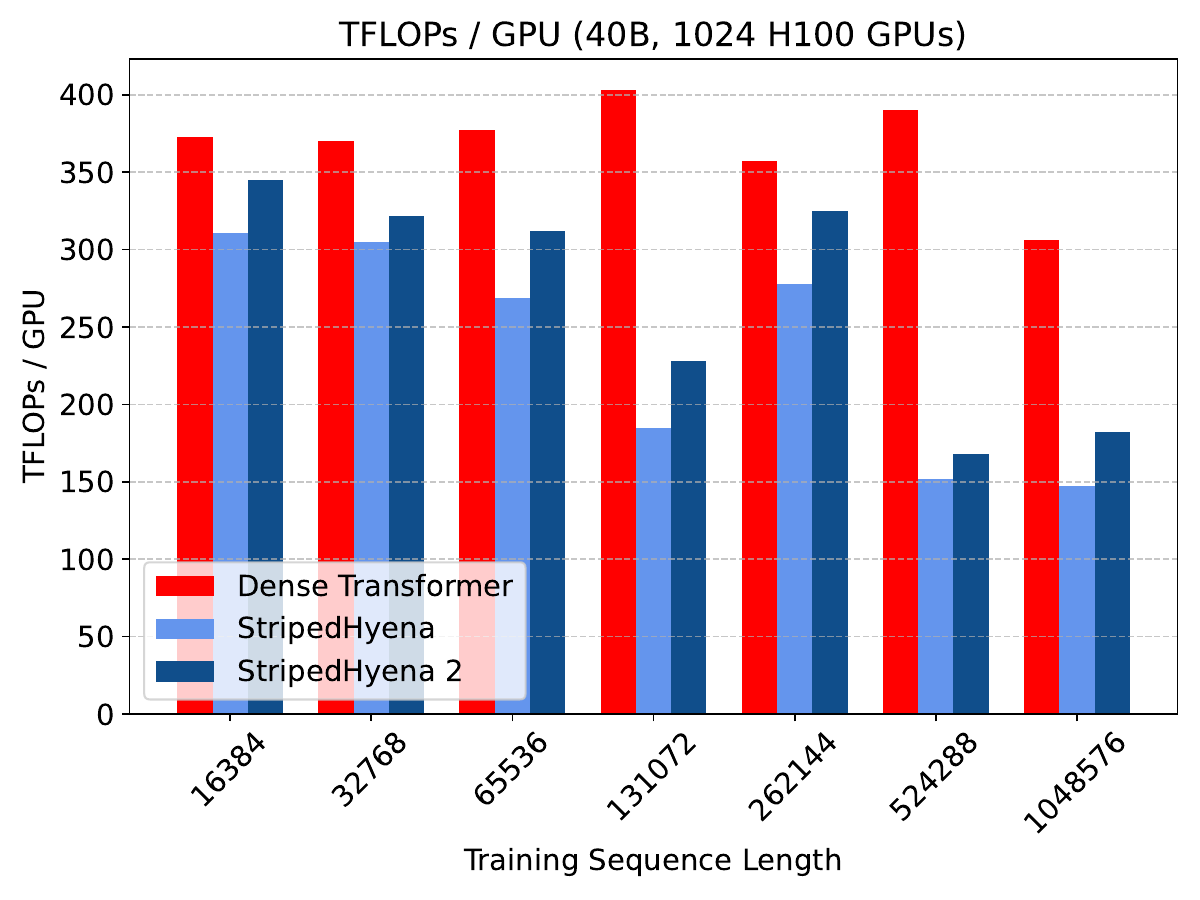} \hfill
    \includegraphics[width=0.45\textwidth]{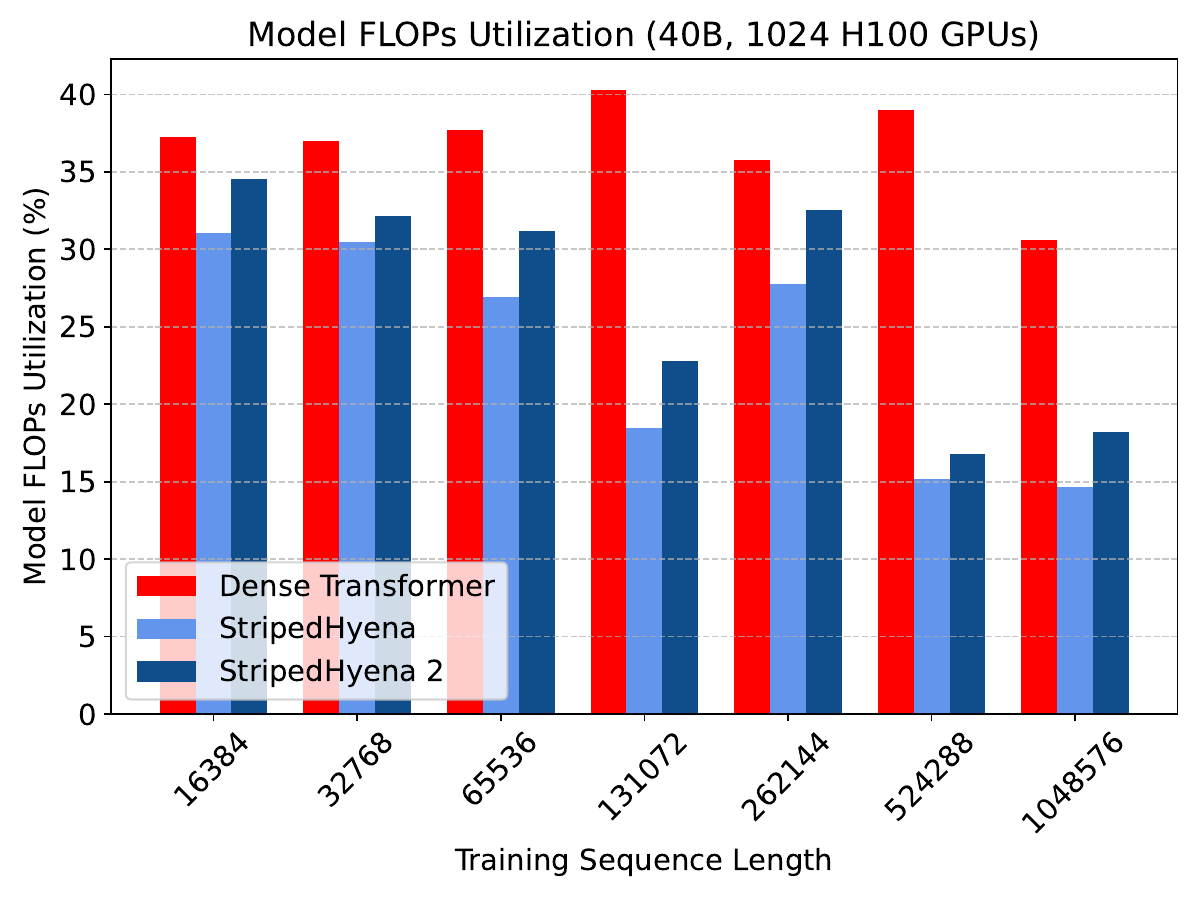}
    \caption{MFU and TFLOPS / s / GPU of 40B models with same distributed configuration and different architectures. For StripedHyena 2, we achieve peak MFU of 34\% at 16K context. See Table \ref{tab:appendix_scaling} for details on the measurement protocol.}
    \label{fig:all_scaling2}
\end{figure}

\begin{figure}[H]
    \centering
    \includegraphics[width=0.8\textwidth]{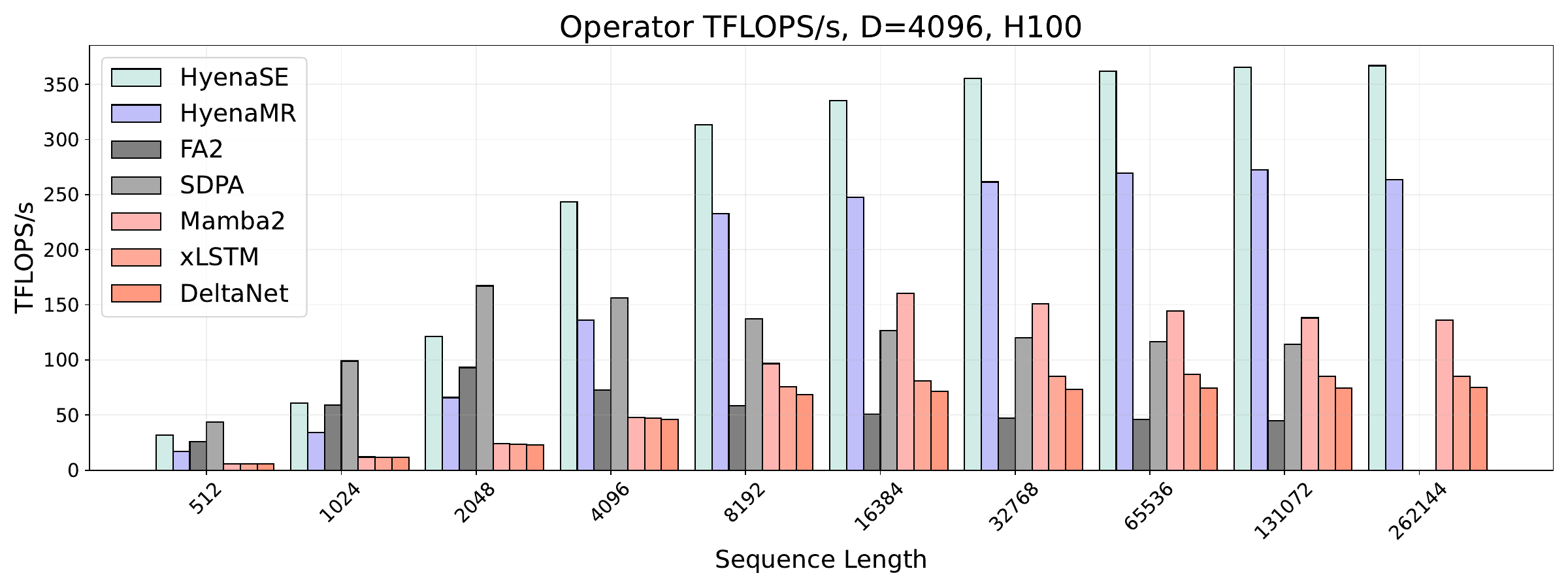}
    \caption{Forward TFLOPs / second of \textsf{Hyena-SE}, \textsf{Hyena-MR} and other common operators in architecture design.}
    \label{fig:tflop-app}
\end{figure}

%% file: sections/appendix/C_protocol.tex
\section{Appendix: Methods}

\begin{table}[H]
    \centering
    \begin{minipage}[t]{0.49\textwidth}
        \centering
        \begin{tabular}{ll}
        \toprule
        \rowcolor{blue!10} \textbf{\textsf{Setting}} & \textbf{\textsf{Value}} \\
        \midrule
        \textsf{Tensor Parallel}  & 2, 2, 8, 8, 16, 16, 32 \bigstrut \\
        \textsf{Sequence Parallel} & True \\
        \textsf{Context Parallel} & 1, 1, 1, 1, 1, 2, 2 \bigstrut \\
        \textsf{Global Batch Size} & 4M tokens \bigstrut \\
        \textsf{Hardware} & H100 SXM \bigstrut \\
        \textsf{GPU Count} & 256 \bigstrut \\
            \bottomrule
        \end{tabular}
    \end{minipage}
    \begin{minipage}[t]{0.49\textwidth}
        \centering
        \begin{tabular}{ll}
        \toprule
        \rowcolor{blue!10} \textbf{\textsf{Setting}} & \textbf{\textsf{Value}} \\
        \midrule
        \textsf{Tensor Parallel} & 8, 8, 8, 8, 16, 32, 64 \bigstrut \\
        \textsf{Sequence Parallel} & True \bigstrut \\
        \textsf{Context Parallel} & 1, 1, 1, 2, 2, 2, 2 \bigstrut \\
        \textsf{Global Batch Size} & 8M tokens \bigstrut \\
        \textsf{Hardware} & H100 SXM \bigstrut \\
        \textsf{GPU Count} & 2048 \bigstrut \\
        \bottomrule
    \end{tabular}
    \end{minipage}
    \caption{\textbf{[Left]:} Baseline distributed configuration used for $7$B parameter model measurements in Figure \ref{fig:all_scaling} at 16K, $33$K, $65$K, $131$K, $262$K, $524$K, $1$M sequence length. 
    \textbf{[Right]:} Baseline distributed training configuration used for $40$B parameter model measurements in Figure \ref{fig:all_scaling} at $16$K, $33$K, $65$K, $131$K, $262$K, $524$K, $1$M sequence length. We also verify scaling on $2048$ with the same settings, doubling batch size $16$M. We observe similar throughput multipliers even on $2048$.}
    \label{tab:appendix_scaling}
\end{table}

\subsection{Pretraining Experiments}\label{sup-sec:grouping}

\paragraph{Methodology}

For all training experiments, we use Savanna. Configuration files are available in the repository, at the following frozen commit hash: \href{github.com/Zymrael/savanna/commit/5f9fdbf1bff7c3ff6e225d46d32d4d7eb97fead9}{{\tt 5f9fdb}}. We train on the {\tt OpenGenome 2} dataset \citep{brixievo2}. For throughput measurements, we report the settings in dedicated tables (Table \ref{tab:appendix_scaling}). We use critical batch size estimation \citep{mccandlish2018empirical} to determine a batch size for the 7 billion parameter runs. We train in mixed precision using FP8 for dense layers and norms.

\paragraph{Effect of grouped convolution}
 
We train 7 billion parameter StripedHyena 2 \textsf{SE-MR-LI} models on 400 billion tokens {\tt OpenGenome2}, with group sizes $1$ (baseline) and $16$, and observe no significant difference in convergence. We also explore the effect of filter grouping in Hyena on smaller models, including simpler hybrids using only \textsf{Hyena-SE} and \textsf{Hyena-MR}. Group sizes larger than $64$ introduce minimal degradation in quality, more visible in smaller models. A smaller number of independent filters also reduces the granularity of the regularization in \textsf{Hyena-MR}, initialized to span different values across filters. The configuration files to replicate the experiments are available in Savanna: \href{https://github.com/Zymrael/savanna/tree/main/configs/7b-final-test/model_configs/group}{{\tt configs/7b-final-test/}}.

\paragraph{Context extension}

Full configuration files are available in \href{https://github.com/Zymrael/savanna/tree/main/configs/context-scaling}{{\tt configs/context-scaling}}. We evaluate recall during midtraining context extension with the needle-in-a-haystack task described in \citep{brixievo2}. Results are shown in Figure \ref{fig:niah_extension}.

\paragraph{Replacing feed-forward-layers with convolutions}

In early designs, we also trained variants of the architecture where every feed-forward layer (MLP, SwiGLU) following every hyena or MHA operator had been replaced with \textsf{Hyena-SE}. We observe generally improved convergence for these models with a small decrease in throughput. Given these findings, coupled with the higher throughput of \textsf{Hyena-SE} compared to MHA or state-space models, we expect future multi-hybrid designs to also optimize the ratio of MLPs and \textsf{Hyena-SE} (or their MoE variants). Configuration files are available at: \href{https://github.com/Zymrael/savanna/tree/main/configs/model/evo2/ablations}{{\tt configs/model/evo2/ablations}}.

\subsection{Profiling}

\paragraph{Operator profiling}

All operators use their official kernels\footnote{We use the mLSTM kernels developed for \href{https://huggingface.co/NX-AI/xLSTM-7b}{xLSTM-7B}.}. For DeltaNet, we report the latency of the kernel provided in the Flash Linear Attention \citep{yang2024fla} repository. For Mamba2 \citep{dao2024transformers}, we use the official kernels provided by the authors.

Figure \ref{fig:operator_profiles} and \ref{fig:tflop-app} shows the results, with \textsf{Hyena-SE} and \textsf{Hyena-MR} providing graceful scaling to longer sequence lengths, with high throughput even at batch size 1. Figure \ref{fig:hyenamr_variants} shows a direct comparison of latencies and TFLOPS / second for the \textsf{Hyena-MR} with different underlying implementations. Direct convolutions using the proposed two-stage approach generally outperform PyTorch convolutions.

%% file: main.bbl
\begin{thebibliography}{56}
\providecommand{\natexlab}[1]{#1}
\providecommand{\url}[1]{\texttt{#1}}
\expandafter\ifx\csname urlstyle\endcsname\relax
  \providecommand{\doi}[1]{doi: #1}\else
  \providecommand{\doi}{doi: \begingroup \urlstyle{rm}\Url}\fi

\bibitem[Ainslie et~al.(2023)Ainslie, Lee-Thorp, de~Jong, Zemlyanskiy, Lebr{\'o}n, and Sanghai]{ainslie2023gqa}
Joshua Ainslie, James Lee-Thorp, Michiel de~Jong, Yury Zemlyanskiy, Federico Lebr{\'o}n, and Sumit Sanghai.
\newblock Gqa: Training generalized multi-query transformer models from multi-head checkpoints.
\newblock \emph{arXiv preprint arXiv:2305.13245}, 2023.

\bibitem[Aky{\"u}rek et~al.(2024)Aky{\"u}rek, Wang, Kim, and Andreas]{akyurek2024context}
Ekin Aky{\"u}rek, Bailin Wang, Yoon Kim, and Jacob Andreas.
\newblock In-context language learning: Arhitectures and algorithms.
\newblock \emph{arXiv preprint arXiv:2401.12973}, 2024.

\bibitem[Arora et~al.(2023)Arora, Eyuboglu, Timalsina, Johnson, Poli, Zou, Rudra, and R{\'e}]{arora2023zoology}
Simran Arora, Sabri Eyuboglu, Aman Timalsina, Isys Johnson, Michael Poli, James Zou, Atri Rudra, and Christopher R{\'e}.
\newblock Zoology: Measuring and improving recall in efficient language models.
\newblock \emph{arXiv preprint arXiv:2312.04927}, 2023.

\bibitem[Beck et~al.(2024)Beck, P{\"o}ppel, Spanring, Auer, Prudnikova, Kopp, Klambauer, Brandstetter, and Hochreiter]{beck2024xlstm}
Maximilian Beck, Korbinian P{\"o}ppel, Markus Spanring, Andreas Auer, Oleksandra Prudnikova, Michael Kopp, G{\"u}nter Klambauer, Johannes Brandstetter, and Sepp Hochreiter.
\newblock xlstm: Extended long short-term memory.
\newblock \emph{arXiv preprint arXiv:2405.04517}, 2024.

\bibitem[Beltagy et~al.(2020)Beltagy, Peters, and Cohan]{beltagy2020longformer}
Iz~Beltagy, Matthew~E Peters, and Arman Cohan.
\newblock Longformer: The long-document transformer.
\newblock \emph{arXiv preprint arXiv:2004.05150}, 2020.

\bibitem[Bouguezel et~al.(2004)Bouguezel, Ahmad, and Swamy]{bouguezel2004improved}
Saad Bouguezel, M~Omair Ahmad, and MNS Swamy.
\newblock An improved radix-16 fft algorithm.
\newblock In \emph{Canadian Conference on Electrical and Computer Engineering 2004 (IEEE Cat. No. 04CH37513)}, volume~2, pp.\  1089--1092. IEEE, 2004.

\bibitem[Brandon et~al.(2023)Brandon, Nrusimha, Qian, Ankner, Jin, Song, and Ragan-Kelley]{brandon2023striped}
William Brandon, Aniruddha Nrusimha, Kevin Qian, Zachary Ankner, Tian Jin, Zhiye Song, and Jonathan Ragan-Kelley.
\newblock Striped attention: Faster ring attention for causal transformers.
\newblock \emph{arXiv preprint arXiv:2311.09431}, 2023.

\bibitem[Brixi et~al.(2025)Brixi, Durrant, Ku, Poli, Brockman, Chang, Gonzalez, King, Li, Merchant, Naghipourfar, Nguyen, Ricci-Tam, Romero, Sun, Taghibakshi, Vorontsov, Yang, Deng, Gorton, Nguyen, Wang, Adams, Baccus, Dillmann, Ermon, Guo, Ilango, Janik, Lu, Mehta, Mofrad, Ng, Pannu, Re, Schmok, St.~John, Sullivan, Zhu, Zynda, Balsam, Collison, Costa, Hernandez-Boussard, Ho, Liu, McGrath, Powell, Burke, Goodarzi, Hsu, and Hie]{brixievo2}
Garyk Brixi, Matthew~G Durrant, Jerome Ku, Michael Poli, Greg Brockman, Daniel Chang, Gabriel~A Gonzalez, Samuel~H King, David~B Li, Aditi~T Merchant, Mohsen Naghipourfar, Eric Nguyen, Chiara Ricci-Tam, David~W Romero, Gwanggyu Sun, Ali Taghibakshi, Anton Vorontsov, Brandon Yang, Myra Deng, Liv Gorton, Nam Nguyen, Nicholas~K Wang, Etowah Adams, Stephen~A Baccus, Steven Dillmann, Stefano Ermon, Daniel Guo, Rajesh Ilango, Ken Janik, Amy~X Lu, Reshma Mehta, Mohammad~R.K. Mofrad, Madelena~Y Ng, Jaspreet Pannu, Christopher Re, Jonathan~C Schmok, John St.~John, Jeremy Sullivan, Kevin Zhu, Greg Zynda, Daniel Balsam, Patrick Collison, Anthony~B. Costa, Tina Hernandez-Boussard, Eric Ho, Ming-Yu Liu, Tom McGrath, Kimberly Powell, Dave~P. Burke, Hani Goodarzi, Patrick~D Hsu, and Brian Hie.
\newblock Genome modeling and design across all domains of life with evo 2.
\newblock \emph{bioRxiv}, 2025.
\newblock \doi{10.1101/2025.02.18.638918}.
\newblock URL \url{https://www.biorxiv.org/content/early/2025/02/21/2025.02.18.638918}.

\bibitem[Brown et~al.(2020)Brown, Mann, Ryder, Subbiah, Kaplan, Dhariwal, Neelakantan, Shyam, Sastry, Askell, et~al.]{brown2020language}
Tom Brown, Benjamin Mann, Nick Ryder, Melanie Subbiah, Jared~D Kaplan, Prafulla Dhariwal, Arvind Neelakantan, Pranav Shyam, Girish Sastry, Amanda Askell, et~al.
\newblock Language models are few-shot learners.
\newblock \emph{Advances in neural information processing systems}, 33:\penalty0 1877--1901, 2020.

\bibitem[Burrus \& Parks(1985)Burrus and Parks]{burrus1985convolution}
C~Sidney Burrus and T~Parks.
\newblock Convolution algorithms.
\newblock \emph{Citeseer: New York, NY, USA}, 6:\penalty0 15, 1985.

\bibitem[Chen et~al.(2023)Chen, Wong, Chen, and Tian]{chen2023extending}
Shouyuan Chen, Sherman Wong, Liangjian Chen, and Yuandong Tian.
\newblock Extending context window of large language models via positional interpolation.
\newblock \emph{URL https://arxiv. org/abs/2306.15595}, 2023.

\bibitem[Chetlur et~al.(2014)Chetlur, Woolley, Vandermersch, Cohen, Tran, Catanzaro, and Shelhamer]{chetlur2014cudnn}
Sharan Chetlur, Cliff Woolley, Philippe Vandermersch, Jonathan Cohen, John Tran, Bryan Catanzaro, and Evan Shelhamer.
\newblock cudnn: Efficient primitives for deep learning.
\newblock \emph{arXiv preprint arXiv:1410.0759}, 2014.

\bibitem[Child et~al.(2019)Child, Gray, Radford, and Sutskever]{child2019generating}
Rewon Child, Scott Gray, Alec Radford, and Ilya Sutskever.
\newblock Generating long sequences with sparse transformers.
\newblock \emph{arXiv preprint arXiv:1904.10509}, 2019.

\bibitem[Dai et~al.(2021)Dai, Liu, Le, and Tan]{dai2021coatnet}
Zihang Dai, Hanxiao Liu, Quoc~V Le, and Mingxing Tan.
\newblock Coatnet: Marrying convolution and attention for all data sizes.
\newblock \emph{Advances in neural information processing systems}, 34:\penalty0 3965--3977, 2021.

\bibitem[Dao(2023)]{dao2023flashattention}
Tri Dao.
\newblock Flashattention-2: Faster attention with better parallelism and work partitioning.
\newblock \emph{arXiv preprint arXiv:2307.08691}, 2023.

\bibitem[Dao \& Gu(2024)Dao and Gu]{dao2024transformers}
Tri Dao and Albert Gu.
\newblock Transformers are ssms: Generalized models and efficient algorithms through structured state space duality.
\newblock \emph{arXiv preprint arXiv:2405.21060}, 2024.

\bibitem[Dubey et~al.(2024)Dubey, Jauhri, Pandey, Kadian, Al-Dahle, Letman, Mathur, Schelten, Yang, Fan, et~al.]{dubey2024llama}
Abhimanyu Dubey, Abhinav Jauhri, Abhinav Pandey, Abhishek Kadian, Ahmad Al-Dahle, Aiesha Letman, Akhil Mathur, Alan Schelten, Amy Yang, Angela Fan, et~al.
\newblock The llama 3 herd of models.
\newblock \emph{arXiv preprint arXiv:2407.21783}, 2024.

\bibitem[Fathi et~al.(2023)Fathi, Pilault, Bacon, Pal, Firat, and Goroshin]{fathi2023block}
Mahan Fathi, Jonathan Pilault, Pierre-Luc Bacon, Christopher Pal, Orhan Firat, and Ross Goroshin.
\newblock Block-state transformer.
\newblock \emph{arXiv preprint arXiv:2306.09539}, 2023.

\bibitem[Fu et~al.(2022)Fu, Dao, Saab, Thomas, Rudra, and R{\'e}]{fu2022hungry}
Daniel~Y Fu, Tri Dao, Khaled~K Saab, Armin~W Thomas, Atri Rudra, and Christopher R{\'e}.
\newblock Hungry hungry hippos: Towards language modeling with state space models.
\newblock \emph{arXiv preprint arXiv:2212.14052}, 2022.

\bibitem[Fu et~al.(2023)Fu, Kumbong, Nguyen, and R{\'e}]{fu2023flashfftconv}
Daniel~Y Fu, Hermann Kumbong, Eric Nguyen, and Christopher R{\'e}.
\newblock Flashfftconv: Efficient convolutions for long sequences with tensor cores.
\newblock \emph{arXiv preprint arXiv:2311.05908}, 2023.

\bibitem[Glorioso et~al.(2024)Glorioso, Anthony, Tokpanov, Whittington, Pilault, Ibrahim, and Millidge]{glorioso2024zamba}
Paolo Glorioso, Quentin Anthony, Yury Tokpanov, James Whittington, Jonathan Pilault, Adam Ibrahim, and Beren Millidge.
\newblock Zamba: A compact 7b ssm hybrid model.
\newblock \emph{arXiv preprint arXiv:2405.16712}, 2024.

\bibitem[Gupta et~al.(2022)Gupta, Gu, and Berant]{gupta2022diagonal}
Ankit Gupta, Albert Gu, and Jonathan Berant.
\newblock Diagonal state spaces are as effective as structured state spaces.
\newblock \emph{Advances in Neural Information Processing Systems}, 35:\penalty0 22982--22994, 2022.

\bibitem[Huang \& Chen(2012)Huang and Chen]{huang2012high}
Shen-Jui Huang and Sau-Gee Chen.
\newblock A high-throughput radix-16 fft processor with parallel and normal input/output ordering for ieee 802.15. 3c systems.
\newblock \emph{IEEE Transactions on Circuits and Systems I: Regular Papers}, 59\penalty0 (8):\penalty0 1752--1765, 2012.

\bibitem[Jacobs et~al.(2023)Jacobs, Tanaka, Zhang, Zhang, Song, Rajbhandari, and He]{jacobs2023deepspeed}
Sam~Ade Jacobs, Masahiro Tanaka, Chengming Zhang, Minjia Zhang, Shuaiwen~Leon Song, Samyam Rajbhandari, and Yuxiong He.
\newblock Deepspeed ulysses: System optimizations for enabling training of extreme long sequence transformer models.
\newblock \emph{arXiv preprint arXiv:2309.14509}, 2023.

\bibitem[Jelassi et~al.(2024)Jelassi, Brandfonbrener, Kakade, and Malach]{jelassi2024repeat}
Samy Jelassi, David Brandfonbrener, Sham~M Kakade, and Eran Malach.
\newblock Repeat after me: Transformers are better than state space models at copying.
\newblock \emph{arXiv preprint arXiv:2402.01032}, 2024.

\bibitem[Katharopoulos et~al.(2020)Katharopoulos, Vyas, Pappas, and Fleuret]{katharopoulos2020transformers}
Angelos Katharopoulos, Apoorv Vyas, Nikolaos Pappas, and Fran{\c{c}}ois Fleuret.
\newblock Transformers are rnns: Fast autoregressive transformers with linear attention.
\newblock In \emph{International conference on machine learning}, pp.\  5156--5165. PMLR, 2020.

\bibitem[Li et~al.(2021)Li, Cheng, and Lin]{li2021tcfft}
Binrui Li, Shenggan Cheng, and James Lin.
\newblock tcfft: Accelerating half-precision fft through tensor cores.
\newblock \emph{arXiv preprint arXiv:2104.11471}, 2021.

\bibitem[Lieber et~al.(2024)Lieber, Lenz, Bata, Cohen, Osin, Dalmedigos, Safahi, Meirom, Belinkov, Shalev-Shwartz, et~al.]{lieber2024jamba}
Opher Lieber, Barak Lenz, Hofit Bata, Gal Cohen, Jhonathan Osin, Itay Dalmedigos, Erez Safahi, Shaked Meirom, Yonatan Belinkov, Shai Shalev-Shwartz, et~al.
\newblock Jamba: A hybrid transformer-mamba language model.
\newblock \emph{arXiv preprint arXiv:2403.19887}, 2024.

\bibitem[Liu et~al.(2024)Liu, Feng, Wang, Wang, Liu, Zhao, Dengr, Ruan, Dai, Guo, et~al.]{liu2024deepseek}
Aixin Liu, Bei Feng, Bin Wang, Bingxuan Wang, Bo~Liu, Chenggang Zhao, Chengqi Dengr, Chong Ruan, Damai Dai, Daya Guo, et~al.
\newblock Deepseek-v2: A strong, economical, and efficient mixture-of-experts language model.
\newblock \emph{arXiv preprint arXiv:2405.04434}, 2024.

\bibitem[Liu et~al.(2023)Liu, Zaharia, and Abbeel]{liu2023ring}
Hao Liu, Matei Zaharia, and Pieter Abbeel.
\newblock Ring attention with blockwise transformers for near-infinite context.
\newblock \emph{arXiv preprint arXiv:2310.01889}, 2023.

\bibitem[Massaroli et~al.(2024)Massaroli, Poli, Fu, Kumbong, Parnichkun, Romero, Timalsina, McIntyre, Chen, Rudra, et~al.]{massaroli2024laughing}
Stefano Massaroli, Michael Poli, Dan Fu, Hermann Kumbong, Rom Parnichkun, David Romero, Aman Timalsina, Quinn McIntyre, Beidi Chen, Atri Rudra, et~al.
\newblock Laughing hyena distillery: Extracting compact recurrences from convolutions.
\newblock \emph{Advances in Neural Information Processing Systems}, 36, 2024.

\bibitem[McCandlish et~al.(2018)McCandlish, Kaplan, Amodei, and Team]{mccandlish2018empirical}
Sam McCandlish, Jared Kaplan, Dario Amodei, and OpenAI~Dota Team.
\newblock An empirical model of large-batch training.
\newblock \emph{arXiv preprint arXiv:1812.06162}, 2018.

\bibitem[Nguyen et~al.(2024)Nguyen, Poli, Durrant, Kang, Katrekar, Li, Bartie, Thomas, King, Brixi, et~al.]{nguyen2024sequence}
Eric Nguyen, Michael Poli, Matthew~G Durrant, Brian Kang, Dhruva Katrekar, David~B Li, Liam~J Bartie, Armin~W Thomas, Samuel~H King, Garyk Brixi, et~al.
\newblock Sequence modeling and design from molecular to genome scale with evo.
\newblock \emph{Science}, 386\penalty0 (6723):\penalty0 eado9336, 2024.

\bibitem[Orvieto et~al.(2023)Orvieto, Smith, Gu, Fernando, Gulcehre, Pascanu, and De]{orvieto2023resurrecting}
Antonio Orvieto, Samuel~L Smith, Albert Gu, Anushan Fernando, Caglar Gulcehre, Razvan Pascanu, and Soham De.
\newblock Resurrecting recurrent neural networks for long sequences.
\newblock In \emph{International Conference on Machine Learning}, pp.\  26670--26698. PMLR, 2023.

\bibitem[Peng et~al.(2023)Peng, Alcaide, Anthony, Albalak, Arcadinho, Biderman, Cao, Cheng, Chung, Grella, et~al.]{peng2023rwkv}
Bo~Peng, Eric Alcaide, Quentin Anthony, Alon Albalak, Samuel Arcadinho, Stella Biderman, Huanqi Cao, Xin Cheng, Michael Chung, Matteo Grella, et~al.
\newblock Rwkv: Reinventing rnns for the transformer era.
\newblock \emph{arXiv preprint arXiv:2305.13048}, 2023.

\bibitem[Poli et~al.(2023{\natexlab{a}})Poli, Massaroli, Nguyen, Fu, Dao, Baccus, Bengio, Ermon, and R{\'e}]{poli2023hyena}
Michael Poli, Stefano Massaroli, Eric Nguyen, Daniel~Y Fu, Tri Dao, Stephen Baccus, Yoshua Bengio, Stefano Ermon, and Christopher R{\'e}.
\newblock Hyena hierarchy: Towards larger convolutional language models.
\newblock In \emph{International Conference on Machine Learning}, pp.\  28043--28078. PMLR, 2023{\natexlab{a}}.

\bibitem[Poli et~al.(2023{\natexlab{b}})Poli, Wang, Massaroli, Quesnelle, Carlow, Nguyen, and Thomas]{polistripedhyena}
Michael Poli, Jue Wang, Stefano Massaroli, Jeffrey Quesnelle, Ryan Carlow, Eric Nguyen, and Armin Thomas.
\newblock Stripedhyena: Moving beyond transformers with hybrid signal processing models, 12 2023b.
\newblock \emph{URL https://github.com/togethercomputer/stripedhyena}, 2023{\natexlab{b}}.

\bibitem[Poli et~al.(2024)Poli, Thomas, Nguyen, Ponnusamy, Deiseroth, Kersting, Suzuki, Hie, Ermon, R{\'e}, et~al.]{poli2024mechanistic}
Michael Poli, Armin~W Thomas, Eric Nguyen, Pragaash Ponnusamy, Bj{\"o}rn Deiseroth, Kristian Kersting, Taiji Suzuki, Brian Hie, Stefano Ermon, Christopher R{\'e}, et~al.
\newblock Mechanistic design and scaling of hybrid architectures.
\newblock \emph{arXiv preprint arXiv:2403.17844}, 2024.

\bibitem[Rajbhandari et~al.(2020)Rajbhandari, Rasley, Ruwase, and He]{rajbhandari2020zero}
Samyam Rajbhandari, Jeff Rasley, Olatunji Ruwase, and Yuxiong He.
\newblock Zero: Memory optimizations toward training trillion parameter models.
\newblock In \emph{SC20: International Conference for High Performance Computing, Networking, Storage and Analysis}, pp.\  1--16. IEEE, 2020.

\bibitem[Romero et~al.(2021)Romero, Kuzina, Bekkers, Tomczak, and Hoogendoorn]{romero2021ckconv}
David~W Romero, Anna Kuzina, Erik~J Bekkers, Jakub~M Tomczak, and Mark Hoogendoorn.
\newblock Ckconv: Continuous kernel convolution for sequential data.
\newblock \emph{arXiv preprint arXiv:2102.02611}, 2021.

\bibitem[Shah et~al.(2024)Shah, Bikshandi, Zhang, Thakkar, Ramani, and Dao]{shah2024flashattention}
Jay Shah, Ganesh Bikshandi, Ying Zhang, Vijay Thakkar, Pradeep Ramani, and Tri Dao.
\newblock Flashattention-3: Fast and accurate attention with asynchrony and low-precision.
\newblock \emph{arXiv preprint arXiv:2407.08608}, 2024.

\bibitem[Shazeer(2019)]{shazeer2019fast}
Noam Shazeer.
\newblock Fast transformer decoding: One write-head is all you need.
\newblock \emph{arXiv preprint arXiv:1911.02150}, 2019.

\bibitem[Shazeer(2020)]{shazeer2020glu}
Noam Shazeer.
\newblock Glu variants improve transformer.
\newblock \emph{arXiv preprint arXiv:2002.05202}, 2020.

\bibitem[Shazeer et~al.(2017)Shazeer, Mirhoseini, Maziarz, Davis, Le, Hinton, and Dean]{shazeer2017outrageously}
Noam Shazeer, Azalia Mirhoseini, Krzysztof Maziarz, Andy Davis, Quoc Le, Geoffrey Hinton, and Jeff Dean.
\newblock Outrageously large neural networks: The sparsely-gated mixture-of-experts layer.
\newblock \emph{arXiv preprint arXiv:1701.06538}, 2017.

\bibitem[Su et~al.(2024)Su, Ahmed, Lu, Pan, Bo, and Liu]{su2024roformer}
Jianlin Su, Murtadha Ahmed, Yu~Lu, Shengfeng Pan, Wen Bo, and Yunfeng Liu.
\newblock Roformer: Enhanced transformer with rotary position embedding.
\newblock \emph{Neurocomputing}, 568:\penalty0 127063, 2024.

\bibitem[Takahashi(2019)]{takahashi2019fast}
Daisuke Takahashi.
\newblock \emph{Fast Fourier transform algorithms for parallel computers}.
\newblock Springer, 2019.

\bibitem[Team et~al.(2024)Team, Lenz, Arazi, Bergman, Manevich, Peleg, Aviram, Almagor, Fridman, Padnos, et~al.]{team2024jamba}
Jamba Team, Barak Lenz, Alan Arazi, Amir Bergman, Avshalom Manevich, Barak Peleg, Ben Aviram, Chen Almagor, Clara Fridman, Dan Padnos, et~al.
\newblock Jamba-1.5: Hybrid transformer-mamba models at scale.
\newblock \emph{arXiv preprint arXiv:2408.12570}, 2024.

\bibitem[Vasudevan et~al.(2017)Vasudevan, Anderson, and Gregg]{vasudevan2017parallel}
Aravind Vasudevan, Andrew Anderson, and David Gregg.
\newblock Parallel multi channel convolution using general matrix multiplication.
\newblock In \emph{2017 IEEE 28th international conference on application-specific systems, architectures and processors (ASAP)}, pp.\  19--24. IEEE, 2017.

\bibitem[Vaswani et~al.(2021)Vaswani, Ramachandran, Srinivas, Parmar, Hechtman, and Shlens]{vaswani2021scaling}
Ashish Vaswani, Prajit Ramachandran, Aravind Srinivas, Niki Parmar, Blake Hechtman, and Jonathon Shlens.
\newblock Scaling local self-attention for parameter efficient visual backbones.
\newblock In \emph{Proceedings of the IEEE/CVF Conference on Computer Vision and Pattern Recognition}, pp.\  12894--12904, 2021.

\bibitem[Xiong et~al.(2020)Xiong, Yang, He, Zheng, Zheng, Xing, Zhang, Lan, Wang, and Liu]{xiong2020layer}
Ruibin Xiong, Yunchang Yang, Di~He, Kai Zheng, Shuxin Zheng, Chen Xing, Huishuai Zhang, Yanyan Lan, Liwei Wang, and Tieyan Liu.
\newblock On layer normalization in the transformer architecture.
\newblock In \emph{International Conference on Machine Learning}, pp.\  10524--10533. PMLR, 2020.

\bibitem[Xiong et~al.(2023)Xiong, Liu, Molybog, Zhang, Bhargava, Hou, Martin, Rungta, Sankararaman, Oguz, et~al.]{xiong2023effective}
Wenhan Xiong, Jingyu Liu, Igor Molybog, Hejia Zhang, Prajjwal Bhargava, Rui Hou, Louis Martin, Rashi Rungta, Karthik~Abinav Sankararaman, Barlas Oguz, et~al.
\newblock Effective long-context scaling of foundation models.
\newblock \emph{arXiv preprint arXiv:2309.16039}, 2023.

\bibitem[Yang \& Zhang(2024)Yang and Zhang]{yang2024fla}
Songlin Yang and Yu~Zhang.
\newblock Fla: A triton-based library for hardware-efficient implementations of linear attention mechanism, January 2024.
\newblock URL \url{https://github.com/fla-org/flash-linear-attention}.

\bibitem[Yang et~al.(2024)Yang, Wang, Zhang, Shen, and Kim]{yang2024parallelizing}
Songlin Yang, Bailin Wang, Yu~Zhang, Yikang Shen, and Yoon Kim.
\newblock Parallelizing linear transformers with the delta rule over sequence length.
\newblock \emph{arXiv preprint arXiv:2406.06484}, 2024.

\bibitem[Yao et~al.(2024)Yao, Jacobs, Tanaka, Ruwase, Shafi, Subramoni, and Panda]{yao2024training}
Jinghan Yao, Sam~Ade Jacobs, Masahiro Tanaka, Olatunji Ruwase, Aamir Shafi, Hari Subramoni, and Dhabaleswar~K Panda.
\newblock Training ultra long context language model with fully pipelined distributed transformer.
\newblock \emph{arXiv preprint arXiv:2408.16978}, 2024.

\bibitem[Zhang \& Sennrich(2019)Zhang and Sennrich]{zhang2019root}
Biao Zhang and Rico Sennrich.
\newblock Root mean square layer normalization.
\newblock \emph{Advances in Neural Information Processing Systems}, 32, 2019.

\bibitem[Zhao et~al.(2023)Zhao, Gu, Varma, Luo, Huang, Xu, Wright, Shojanazeri, Ott, Shleifer, et~al.]{zhao2023pytorch}
Yanli Zhao, Andrew Gu, Rohan Varma, Liang Luo, Chien-Chin Huang, Min Xu, Less Wright, Hamid Shojanazeri, Myle Ott, Sam Shleifer, et~al.
\newblock Pytorch fsdp: experiences on scaling fully sharded data parallel.
\newblock \emph{arXiv preprint arXiv:2304.11277}, 2023.

\end{thebibliography}
